\documentclass{article}

\PassOptionsToPackage{numbers, compress}{natbib}

    \usepackage[preprint]{neurips_2025}

\usepackage[utf8]{inputenc} 
\usepackage[T1]{fontenc}    
\usepackage{hyperref}       
\usepackage{url}            
\usepackage{booktabs}       
\usepackage{amsfonts}       
\usepackage{nicefrac}       
\usepackage{microtype}      
\usepackage{xcolor}         

\usepackage{makecell}
\usepackage[small]{caption}
\usepackage{graphicx}
\usepackage{amsmath}
\usepackage{booktabs}
\usepackage{multirow}
\usepackage{amsthm}
\usepackage{verbatim}
\usepackage{amsfonts,  amssymb,  mathrsfs}
\usepackage[subfigure]{tocloft}
\usepackage{subfigure}
\usepackage{xcolor}
\usepackage{wrapfig}

\title{Safety2Drive: Safety-Critical Scenario Benchmark for the Evaluation of  Autonomous Driving}

\author{
  Jingzheng Li \\
  Zhongguancun Laboratory\\
  \texttt{maxlijingzheng@163.com} \\
  \And
  Tiancheng Wang \\
  Beihang University \\
  \texttt{tcwang@buaa.edu.cn} \\
  \And
  Xingyu Peng \\
   Beihang University \\
  \texttt{xypeng@buaa.edu.cn} \\
  \And
  Jiacheng Chen \\
   Beihang University \\
  \texttt{jc2001@buaa.edu.cn} \\
  \And
  Zhijun Chen \\
   Beihang University \\
  \texttt{zhijunchen@buaa.edu.cn} \\
  \And
  Bing Li \\
  A*STAR \\
  \texttt{libingsy@buaa.edu.cn} \\
  \And
  Xianglong Liu\thanks{Corresponding author.} \\
   Beihang University \\
  \texttt{xlliu@buaa.edu.cn} \\
}

\begin{document}

\maketitle

\begin{abstract}
Autonomous Driving (AD) systems demand the high levels of safety assurance. Despite significant advancements in AD demonstrated on open-source benchmarks like Longest6 and Bench2Drive, existing datasets still lack regulatory-compliant scenario libraries for closed-loop testing to comprehensively evaluate the functional safety of AD.
Meanwhile, real-world AD accidents often occur in ‌long-tail edge scenarios‌, i.e., safety-critical scenario, that are underrepresented in current driving datasets. 
This scarcity leads to inadequate evaluation of AD performance, posing risks to safety validation and practical deployment.
To address these challenges, we propose Safety2Drive, a safety-critical scenario library designed to evaluate AD systems. Safety2Drive offers three key contributions. 
(1) Safety2Drive comprehensively covers the test items required by standard regulations and contains 70 AD function test items.
(2) Safety2Drive supports the safety-critical scenario generalization. It has the ability to inject safety threats such as natural environment corruptions and adversarial attacks cross camera and LiDAR sensors. (3) Safety2Drive supports multi-dimensional evaluation. In addition to the evaluation of AD systems, it also supports the evaluation of various perception tasks, such as object detection and lane detection.
Safety2Drive provides a paradigm from scenario construction to validation, establishing a standardized test framework for the safe deployment of AD.


\end{abstract}

\section{Introduction}

The field of autonomous driving (AD) has witnessed tremendous advancements,   with three primary approaches dominating the landscape: modular-based~\cite{liang2020pnpnet},   end-to-end~\cite{jia2023think,  chitta2022transfuser} and foundation model-based AD systems~\cite{mei2024continuously}.
Traditional modular-based AD systems such as Autoware~\cite{kato2018autoware} and Apollo~\cite{kochanthara2024safety} are comprised of distinct perception~\cite{liu2023bevfusion},   planning~\cite{li2024think2drive},   and control modules.
The decoupled design between system
components may lead to key information loss during transitions and potentially redundant computation as well. 
In contrast,   end-to-end AD systems map raw sensor inputs directly to control signals through deep neural networks.
These two AD systems as data-driven paradigms depend on extensive and diverse training data,   which can result in shallow semantic comprehension and latent decision errors under complex driving scenarios.
Such limitations motivate the exploration of knowledge-aware AD systems that integrate foundation models equipped with scene understanding and reasoning capabilities,   as exemplified by recent LLM- and VLM-based AD systems~\cite{renz2024carllava,  wang2024omnidrive}.
‌

Despite significant progress in AD under natural driving scenarios, real-world accidents predominantly occur in ‌long-tail corner cases,   such as jaywalking pedestrians causing planning failures~\cite{li2022coda}.
These safety-critical scenarios remain insufficiently captured‌ in publicly available driving datasets,   leading to ‌incomplete evaluation of AD systems.
Specifically,   current validation paradigms exhibit two distinct yet complementary limitations.
The first paradigm predominantly leverages ‌open-loop benchmarks‌, e.g.,   KITTI~\cite{geiger2012we},   nuScenes~\cite{caesar2020nuscenes} and Waymo. However,   as previously analyzed~\cite{zhai2023rethinking},   these datasets suffer from skewed scenario distribution. For instance, 75\% of nuScenes validation set 
consist of non-interactive, straight-line driving scenarios, highlighting the lack of diverse test cases.
The second paradigm employs closed-loop evaluation frameworks such as CARLA on the benchmarks with multi routes,   e.g.,   Town05Long,   Longest6 and Leaderboard V2.
Such benchmarks predominantly feature ‌over-simplified traffic dynamics‌,   failing to examine driving ability of AD systems under complicated and interactive traffic.
To address this limitation,   SafeBench~\cite{xu2022safebench} proposes a scenario-centric validation framework for AD safety assessment. 
However, it only incorporates 8 predefined safety-critical scenarios.
Further, Bench2Drive~\cite{jiabench2drive} enhances scenario diversity for training and evaluation, but its testing suite remains restricted to 220 short-duration scenarios.
This highlights the critical need for comprehensive test scenario capable of addressing multi-dimensional functional safety requirements.

\begin{figure*}[t]
  \centering
\includegraphics[width=0.99\linewidth,  scale=1]{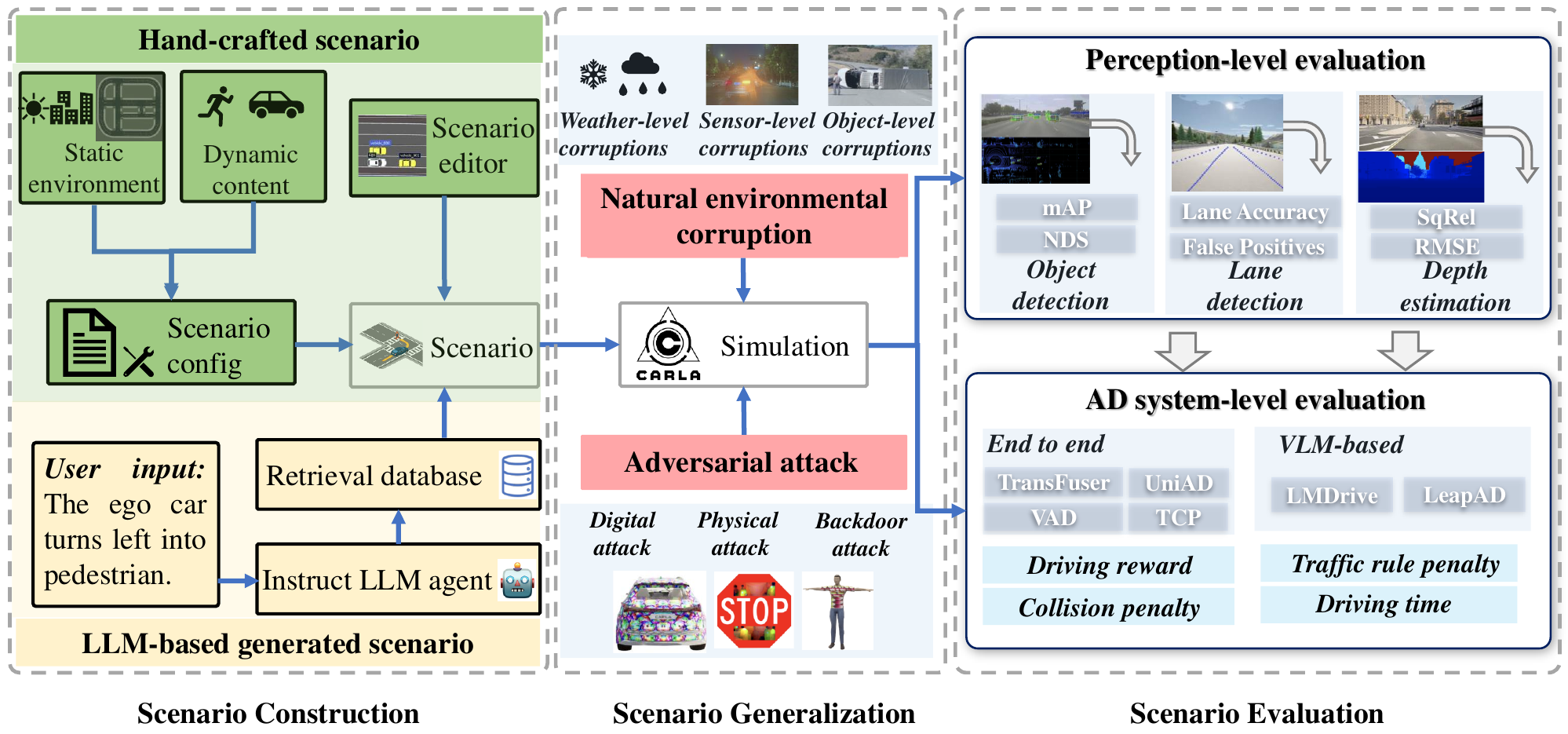}
  \caption{The pipeline of Safety2Drive.}
  \label{motivation}
\end{figure*}

We propose Safety2Drive, a safety-critical benchmark for the evaluation of autonomous driving. 
Fig.\ref{motivation} illustrates the overall test framework of the scenario library, which contains three main modules: scenario construction, scenario generalization and scenario evaluation.
The \textbf{scenario construction} not only includes a library of hand-designed scenarios but also supports LLM-based scenario generation~\cite{zhang2024chatscene} as a flexible supplement.
Safety2Drive is regulatory-compliant scenario library, which comprises 70 functional test items covering most of standardized scenarios mandated by AD standards and regulations such as ISO 34504.
All scenarios are implemented using standardized OpenSCENARIO format with parametric configuration capabilities, allowing dynamic configuration of environmental/traffic parameters for batch testing~\cite{li2024iss}.
The \textbf{scenario generalization} module supports two safety-critical capabilities that are used to transform standard scenarios into safety-critical scenarios: One is natural environmental corruption including weather-level corruption, sensor-level corruptions and object-level corruptions across camera and LiDAR  sensors~\cite{dong2023benchmarking}.
The other is adversarial attack containing three attack vectors including digital attack, physical attack and backdoor attack~\cite{wang2024revisiting}.
The \textbf{scenario evaluation} module establishes a comprehensive evaluation framework spanning from perception tasks to system-level assessment. It not only supports the evaluation of intelligent perception algorithms but also the evaluation of various AD systems.

In summary, the proposed Safety2Drive has the following contributions:

\begin{itemize}
    \item \textbf{Comprehensive standard regulatory functional test coverage.} Safety2Drive includes 70 functional test items. To our knowledge, apart from deploying standard regulatory scenarios in real-world, there is currently no open-source scenario library in closed-loop simulation environments that meet standard regulatory requirements.
    \item \textbf{Safety-critical scenario generalization.} Safety2Drive supports scenario generalization, allowing each functional test item to be expanded into various safety-critical scenarios. Specifically, it supports the injection of safety threats such as natural environment corruptions and adversarial attacks, which can be used to assess the robustness and safety of AD.
    \item \textbf{Evaluation process from perception to system.} Safety2Drive not only supports the evaluation of intelligent perception algorithms but also the evaluation of various AD systems,   realizing a unified component-to-system evaluation framework.
\end{itemize}
\section{Related Work}
\subsection{Autonomous Driving Scenario Benchmarks} 
Benchmarking in the field of AD has seen significant advancements, evolving from specialized datasets tailored for specific tasks to more integrated forms capable of evaluating multiple synergistic system components.
KITTI~\cite{geiger2012we} and CODA~\cite{li2022coda} focus primarily on perception tasks, while datasets like LanEvil~\cite{zhang2024lanevil} aims at lane detection and Trajnet++~\cite{kothari2021human} is used for trajectory prediction.
Further, nuScenes~\cite{caesar2020nuscenes} and Waymo~\cite{sun2020scalability} support the assessment of planning capabilities facilitating the evaluation of various synergistic system components.
Although afore-mentioned datasets provide an open-loop assessment, it is difficult to adequately evaluate planning proficiency due to the lack of closed-loop simulation.
Thus, Longest6, CARLA Leaderboard V2, Bench2Drive~\cite{jiabench2drive} and ISS-Scenario~\cite{li2024iss} provide a closed-loop evaluation with the capability to test AD systems such as \textit{Pedestrian crossing}, \textit{Cut in}, and so on.
Actually, safety-critical scenario is crucial for testing AD systems~\cite{feng2023dense,hao2023adversarial}, but these datasets do not include safety-critical scenario generation capability.
Therefore, we propose a benchmark Safety2Drive that not only has comprehensive functional test items required by standard regulations, but also supports safety-critical scenario generation capability including the natural environmental
corruption and adversarial attack.
Tab.\ref{tab1} presents a comparison of AD scenario benchmarks.
\begin{table*}[t]
\caption{Comparison of autonomous driving scenario benchmarks}
\centering
{
\resizebox{\linewidth}{!}{\begin{tabular}{lccccccc}
\hline
\multirow{2}{*}{Benchmark} & \multirow{2}{*}{Perception }& \multicolumn{3}{c}{Adversarial Attack} & \multirow{2}{*}{Environmental Corruption } & \multirow{2}{*}{Closed-Loop}&\multirow{2}{*}{AD systems}\\
&&Digital Attack&Physical Attack&Backdoor Attack  & &\\
\hline
KITTI~\cite{geiger2012we} & $\times$&$\times$&$\times$&$\times$& $\times$&$\times$&$\times$\\
nuScenes~\cite{caesar2020nuscenes} & $\checkmark$&$\times$&$\times$&$\times$& $\times$&$\times$&$\checkmark$\\
Waymo~\cite{sun2020scalability} & $\checkmark$&$\times$&$\times$&$\times$& $\times$&$\times$&$\checkmark$\\

CommonRoad~\cite{althoff2017commonroad}& $\times$&$\times$&$\times$&$\times$&$\times$& $\checkmark$&$\checkmark$\\
Bench2Drive~\cite{jiabench2drive}& $\checkmark$&$\times$&$\times$&$\times$&$\times$& $\checkmark$&$\checkmark$\\
ISS-Scenario~\cite{li2024iss}&$\times$&$\times$&$\times$&$\times$&$\times$& $\checkmark$&$\checkmark$\\
\hline
Safety2Drive& $\checkmark$&$\checkmark$&$\checkmark$&$\checkmark$& $\checkmark$&$\checkmark$&$\checkmark$\\
\hline
\end{tabular}}
}
\label{tab1}
\end{table*}
\subsection{Evaluation of Autonomous Driving}
Current research focuses on evaluating the safety of autonomous driving from various dimensions, including perception, planning and controller modules~\cite{xu2022safebench,shen2022sok}.
These efforts can be broadly categorized into component-level and system-level evaluation.
The evaluation of perception tasks mainly includes traffic sign recognition~\cite{wang2021can}, lane detection~\cite{zhang2024towards}, and trajectory prediction~\cite{cao2022advdo}.
Besides, some works~\cite{dong2023benchmarking,li2024r} explore the robustness of perception tasks under natural environment corruption and domain shift, which are cases that have a high likelihood to occur in real-world.
For the system-level evaluation, the evaluation metrics typically consider the functional safety, i.e., the collision rate and route completion.
Specifically, PASS~\cite{hu2022pass} is a system-driving evaluation platform for AD security research which analyzes vulnerabilities of AD systems under adversarial attacks.
Further, SafeBench~\cite{xu2022safebench} systematically evaluates the performance of ADs over accident-prone scenarios generated by scenario generation algorithms. 
Safety2Drive includes scenario evaluation module which not only supports the evaluation of perception tasks but also various AD systems.

\section{Safety2Drive}
Central to Safety2Drive is the scenario library.
First we introduce the construction of Safety2Drive including a library of hand-designed scenarios and LLM-based scenario generation ability.
Then we present the scenario generalization module to support safety-critical scenario generalization capabilities such as natural environment corruptions and adversarial attacks.
\subsection{Scenario Construction}
We first conduct a systematic review of current AD function testing regulations issued by relevant governmental authorities,   such as ISO34504,   C-NCAP and E-NCAP.
Subsequently, we summary 6 main categories covering 70 standardized regulatory-compliant functional test items with  parameter configuration, which is summarized in Appendix.
Fig.\ref{visual} presents the visualizations of 4 scenarios alongside their corresponding bird's-eye views (BEV). 
Implementationally, we provide two ways to construct scenario, one is manually designed scenario and the other is LLM-based scenario generation.
Note that the Safety2Drive is hand-crafted due to its accuracy.
LLM-based scenario generation is used as an auxiliary capability for the flexible and fast construction.

\textbf{Hand-crafted scenario.} For each test item, we develop dynamic interaction scenarios in compliance with ASAM OpenSCENARIO standard~\footnote{https://www.asam.net/standards/detail/openscenario/v200/}, which involve configuring behavioral logic of traffic participants (e.g.,   vehicles,   pedestrians),   road conditions, and environmental variables to ensure the functional testing of AD systems.
\begin{wrapfigure}{r}{0.6\textwidth}
  \begin{center}
\includegraphics[width=0.48\textwidth]{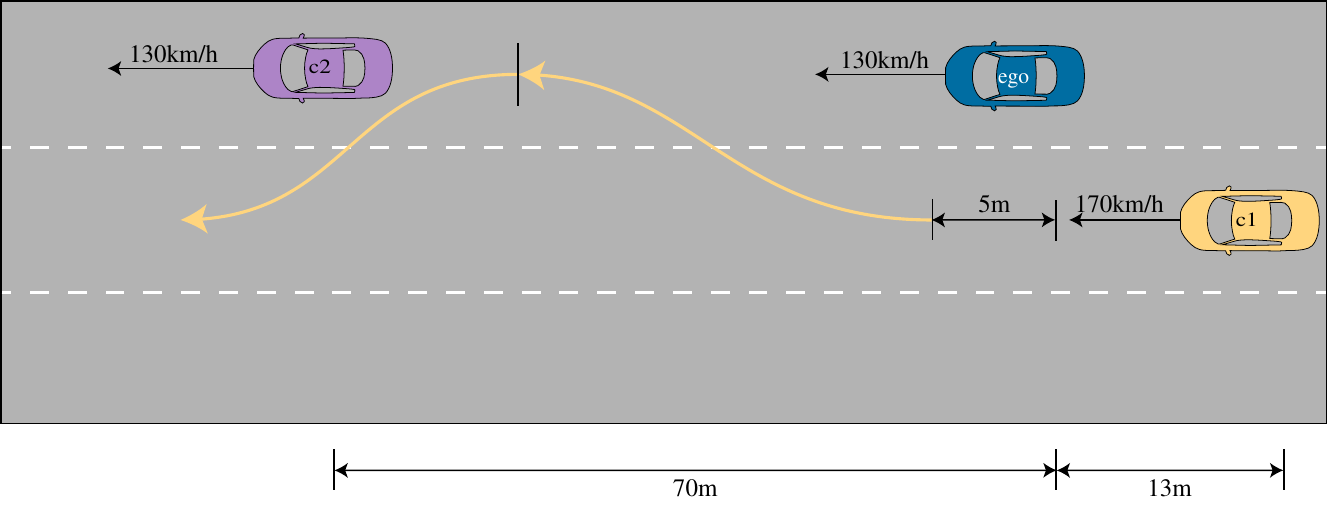}
  \end{center}
  \caption{Example of\textit{ double lane changer} scenario.}
  \label{example1}
\end{wrapfigure}
Fig.\ref{example1} demonstrates an example of scenario description during the creation of a \textit{double lane changer} scenario: 
``the ego vehicle is driving at the rightmost lane behind another vehicle driving at the same speed,   leaving a gap.
A faster vehicle approaches the ego vehicle from behind on the centermost lane. This vehicle changes lane into the gap on the rightmost lane after it has passed the ego vehicle. In order to avoid collision with the vehicle driving ahead of the ego vehicle,   it immediately changes back to the center lane.'' 
Based on the scenario configuration, we construct these scenarios manually in traffic generation editor such as QGIS.

\textbf{LLM-based scenario generation.} 
We adopt the ChatScene framework~\cite{zhang2024chatscene} to generate driving scenarios by interrogating the LLM using scenario descriptions corresponding to functional test items. Implementationally, the LLM model-based scenario generation consists of three key steps.
The first is \textit{user input}: the user begins by providing unstructured language instructions to the LLM agent. 
The second is \textit{instruct LLM agent}: LLM converts user instructions into structured textual descriptions, decomposing the overall scenario into sub-descriptions of specific object behaviors, positions, and interactions.
The third is \textit{retrieval database}: embedding vectors are used to retrieve a pre-built database, matching Scenic code snippets that align with the descriptions. These snippets are then assembled into executable scripts to simulate the scenarios in the CARLA simulation environment.
\subsection{Safety-critical Scenario Generalization}
Based on the foundational functional test items,   Safety2Drive supports safety-critical scenario generalization.
Specifically,   we develop a toolbox capable of extrapolating standardized scenarios into various safety-critical scenarios.
The safety-critical scenario encompasses two primary categories: ‌natural environmental corruptions and adversarial attacks.
The middle of Fig.~\ref{motivation} illustrates the risk element taxonomy integrated within the safety-critical scenario generalization framework.
\subsubsection{Natural Environmental Corruption Scenario}
Natural environmental corruption refers to various distortions and perturbations that can occur in sensor input,   provoking concerns about the safety and
robustness of AD systems.
Following previous works~\cite{beemelmanns2024multicorrupt,  li2024r,  xie2025benchmarking},   we systematically categorize the corruptions into weather-,   sensor- and object-levels.
We then identify common corruption types
for each level considering real-world driving scenarios,   resulting in 20 distinct corruptions in total with 5 severities.
The top of Fig.\ref{attack} shows the visualization of different environmental corruptions.

%

\textbf{Weather-level corruptions }frequently occur in AD scenarios,  degrading the sensing performance of both LiDAR and camera systems.
To systematically evaluate robustness against weather variations,   we focus on 4 common weather-level corruptions: ‌\textit{Snow‌},   ‌\textit{Rain‌},   ‌\textit{Fog‌},   and ‌\textit{Strong Sunlight‌}.

\textbf{Sensor-level corruptions} arise from internal/external factors like vibration,   lighting variations,   and material reflectivity.
Following prior discussions on sensor noises~\cite{dong2023benchmarking},   we adopt 6 corruption types,   three of which are both for the camera and LiDAR sensors,   \textit{Gaussian Noise},   \textit{Uniform Noise},   and \textit{Impulse Noise}.
For the LiDAR sensor,   the point cloud corruptions additionally include \textit{Density Decrease},   \textit{Cutout} and \textit{LiDAR Crosstalk}.

\textbf{Object-level corruptions} refer to the diversity of shapes and states,   viewing angles of the object itself,   making deep learning models difficult to correctly recognize it.
We present 6 main corruptions: \textit{Motion Blur},   \textit{Local Density Decrease},   \textit{Local Cutout},   \textit{Local Gaussian Noise},   \textit{Local Uniform Noise} and \textit{Local Impluse Noise}.
The first corruption is only applied to vision-based objects to simulate the distortions,   which is caused by driving too fast.
The last 5 corruptions are only applied to LiDAR point clouds to simulate the distortions caused by different object materials or occlusion.
\subsubsection{Adversarial Attack Scenario}
\begin{figure*}[t]
  \centering
\includegraphics[width=0.95\linewidth,  scale=1]{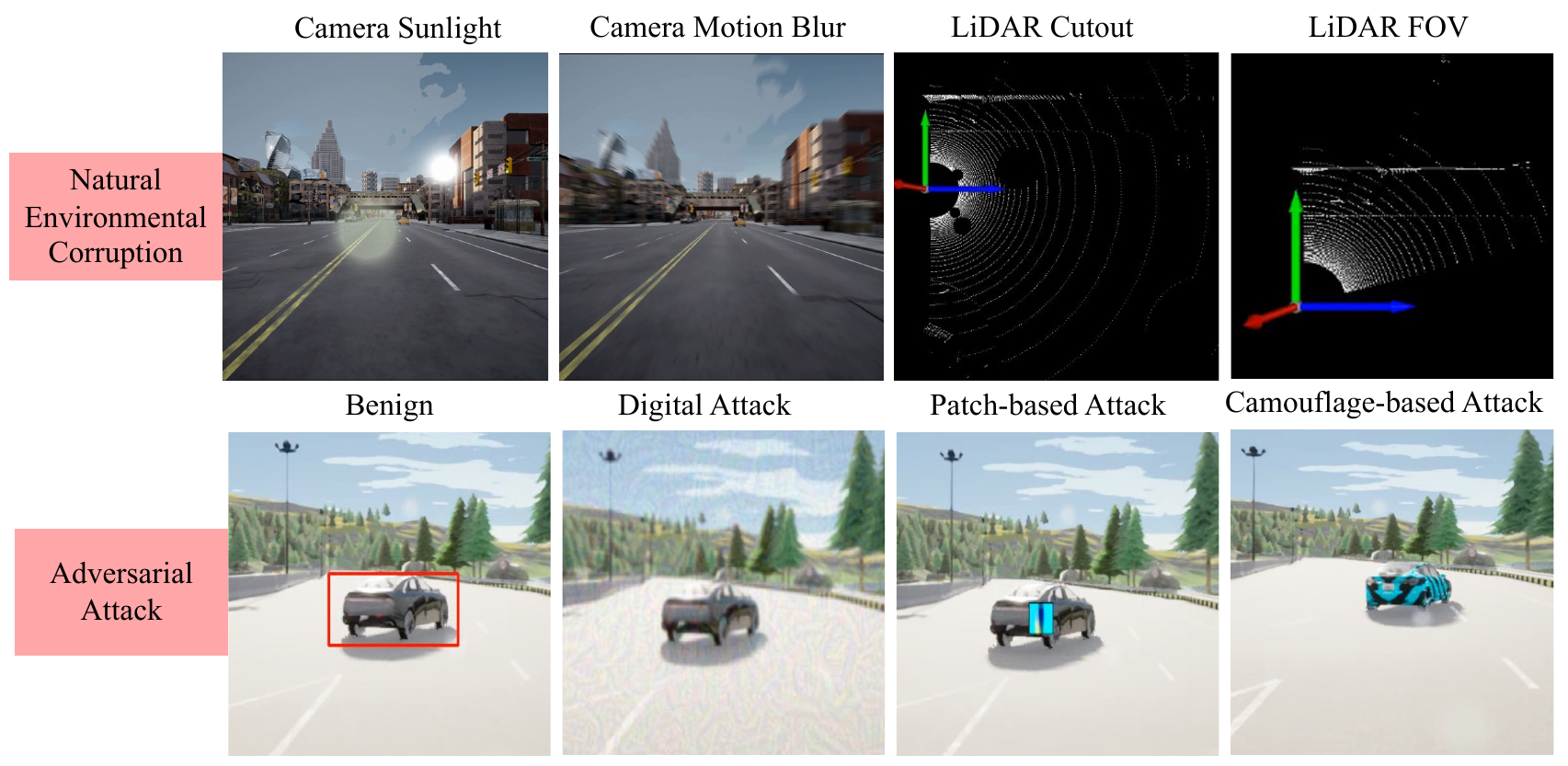}
  \caption{The visualization of safety-critical scenario generalization. The top is natural environmental corruption and the bottom is adversarial attack.}
  \label{attack}
\end{figure*}

In addition to the ability to add natural environmental corruption to AD scenarios,   Safety2Drive also supports adversarial attack scenario generalization by deploying adversarial attack algorithms.
Typical types of adversarial attack include digital attacks,   physical attacks,   and backdoor attacks.
The bottom of Fig.\ref{attack} shows the recognition results of the perception models before and after deploying the adversarial attack algorithms.
\begin{figure*}[t]
  \centering
\includegraphics[width=0.95\linewidth,  scale=1]{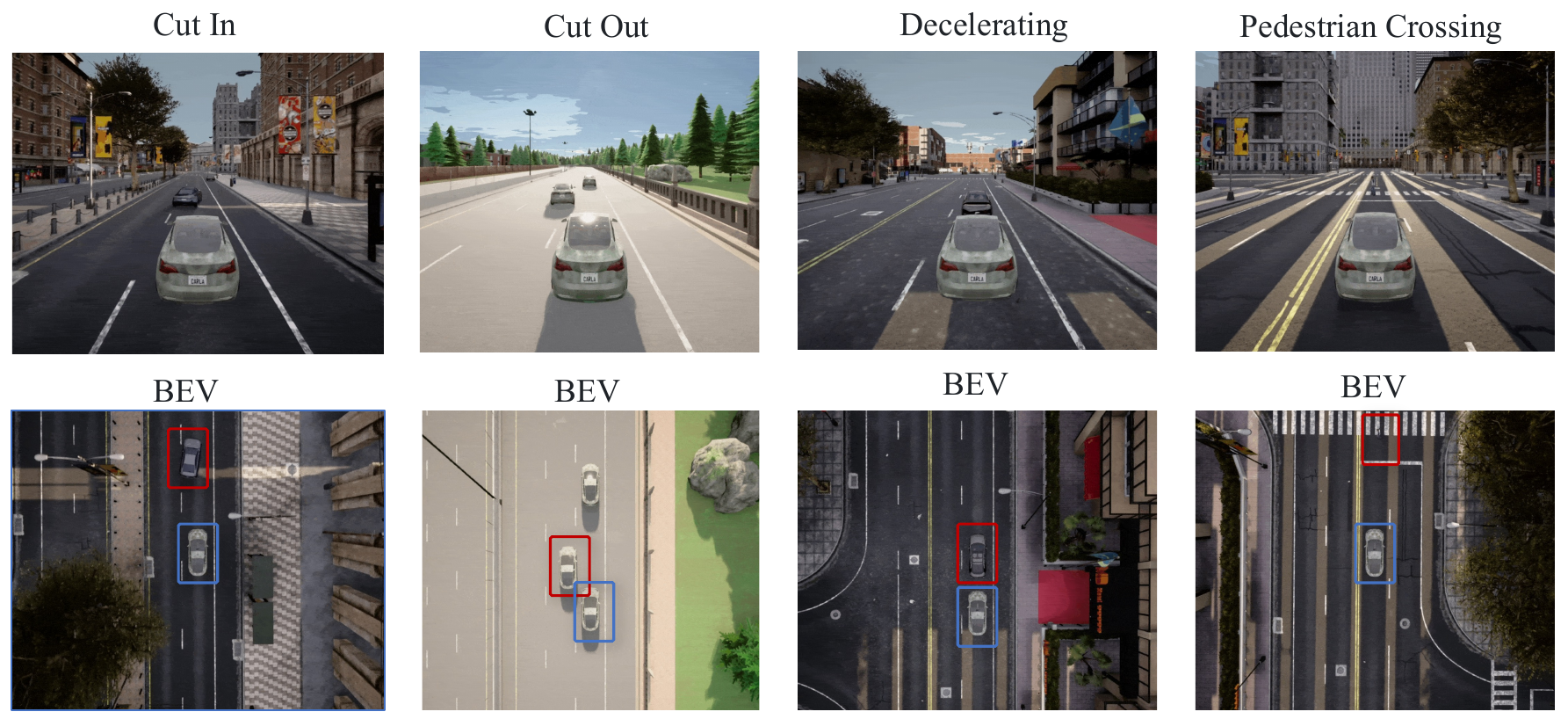}
  \caption{Visualization of four scenarios: \textit{Cut In},   \textit{Cut Out},   \textit{Decelerating},   and \textit{Pedestrian Crossing},   as well as the corresponding BEV. The blue box is the Ego vehicle and the surrounding vehicle and pedestrian are in red.}
  \label{visual}
\end{figure*}
\begin{figure*}[t]
  \centering
\includegraphics[width=0.95\linewidth,  scale=1]{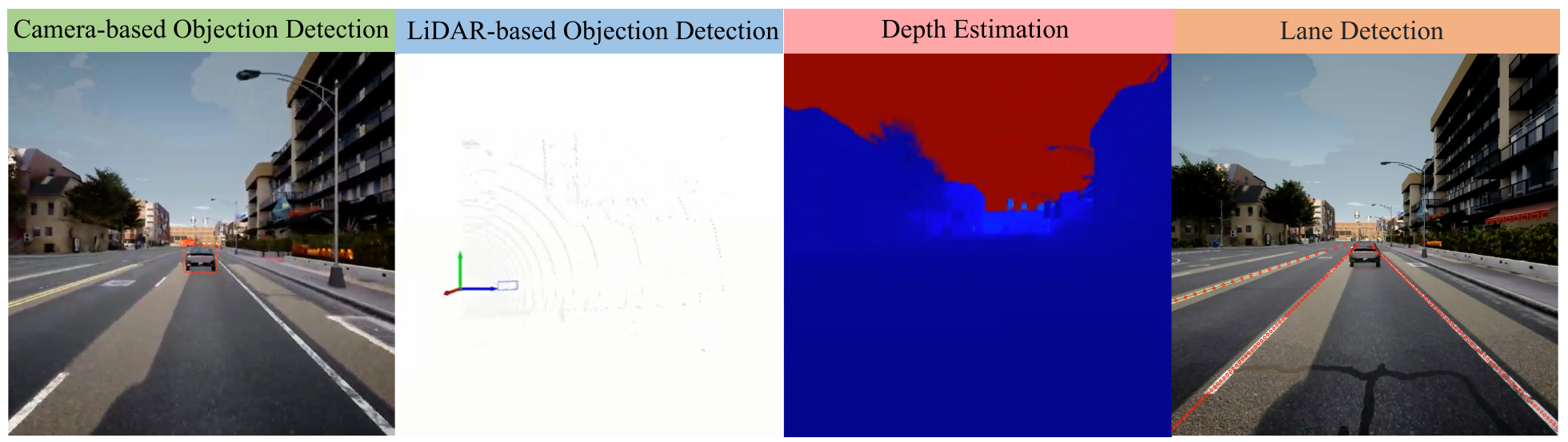}
  \caption{The visualization of prediction results for different perception tasks.}
  \label{perception}
\end{figure*}

\textbf{Digital adversarial attacks.}
Digital attacks target vulnerabilities in neural network models by injecting carefully crafted perturbations to digital inputs,   such as camera images or LiDAR point clouds.
These perturbations maintain visual fidelity imperceptible to human observers while inducing erroneous outputs in perception systems.
State-of-the-art attack methodologies typically leverage gradient-based optimization frameworks,   including methods like Fast Gradient Sign Method (FGSM)~\cite{goodfellow2014explaining} and Projected Gradient Descent (PGD)~\cite{madry2018towards}.

\textbf{Physical adversarial attacks.}
Digital adversarial instances with pixel perturbations are infeasible for physical attack due to pixel-wise modifications.
In practice,   the adversarial attacks~\cite{brown2017adversarial} often need to be deployed in the real world. 
However,   adversarial attacks are applied to 3D objects in the physical world to significantly degrade the detection score of that object in a variety of transformations,   such
as object materials,   camera poses,   lighting conditions,   and
background interactions.
Thus,   the realization of physical adversarial attacks~\cite{liu2024beware} needs to consider digital-to-physical transformation to improve the robustness of adversarial attacks, mainly including expectation over transformation (EOT)~\cite{athalye2018synthesizing},   non-printability score (NPS)~\cite{sharif2016accessorize},   total variant (TV) loss~\cite{mahendran2015understanding},   etc.
The physical adversarial attacks can be grouped into
\textbf{patch-based attack}~\cite{shrestha2023towards} and \textbf{camouflage-based attack}~\cite{wang2021dual,  zhang2023boosting} according to the form
of perturbation.


\textbf{Backdoor attack:}
For backdoor attack~\cite{zhang2024backdoor},   the adversary crafts or learns a trigger embedding function $\boldsymbol{x}^{tri} = \text{Trigger}(\boldsymbol{x})$ that can convert a clean data $\boldsymbol{x}_{i}$ into a triggered/poisoned data $\boldsymbol{x}^{tri}_{i}$. Given
a target label $y^{\prime}$,   the adversary can poison a small part of the training
samples,   i.e.,   replace $(\boldsymbol{x}_i ,   y_i^{\prime})$ with $(\boldsymbol{x}_i^{tri} ,   y_i^{\prime})$,   which produces
poisoned training data $D_{\text {train }}^{\prime}$. 
Training with $D_{\text {train }}^{\prime}$ results in
the infected model $f(\boldsymbol{x};\boldsymbol{\theta})$.
At test time,   if a clean input data $(\boldsymbol{x},   y) \in D_{\text {test }}$ is fed into the infected model,   it should be correctly predicted as $y$. However,   for a triggered data  $\boldsymbol{x}^{tri}_{i}$,   its prediction changes
to the target label $y^{\prime}$.
Backdoor attacks in AD systems create latent vulnerabilities where adversarial triggers can override perception systems intentionally by inducing targeted misbehaviors~\cite{pourkeshavarz2024adversarial}.


\begin{table*}[t]
\Large
\caption{The results of object detection on Safety2Drive with different types of environment corruption.}
\centering
{\resizebox{\linewidth}{!}{
\begin{tabular}{cc|cc|cc|cc|cc|cc|cc}
\toprule
\multicolumn{2}{c}{\multirow{2}{*}{Corruptions}}&\multicolumn{2}{c}{Sudden Pede. Crossing}&\multicolumn{2}{c}{Lane Changing}&\multicolumn{2}{c}{Stationary Obstacle}&\multicolumn{2}{c}{Opposing Passing}&\multicolumn{2}{c}{Cut In}&\multicolumn{2}{c}{Cut Out}\\
&&YoloV5&PointPillars&YoloV5&PointPillars&YoloV5&PointPillars&YoloV5&PointPillars&YoloV5&PointPillars&YoloV5&PointPillars\\
\midrule
\multicolumn{2}{c}{Clean}&0.883&0.621&0.882&0.452&0.770&0.364&0.839&0.516&0.880&0.905&0.924&0.909\\
\midrule
\multirow{4}{*}{Weather}&Snow&0.072&0.621&0.326&0.312&0.113&0.273&0.149&0.426&0.370&0.411&0.158&0.908\\
&Rain&0.306&0.623&0.456&0.076&0.087&0.091&0.511&0.517&0.537&0.318&0.540&0.906\\
&Fog&0.285&0.254&0.647&0.271&0.266&0.091&0.363&0.175&0.532&0.091&0.558&0.636\\
&Sunlight&0.429&0.539&0.638&0.080&0.162&0.268&0.598&0.774&0.673&0.532&0.505&0.358\\
\midrule
\multirow{6}{*}{Sensor}&Gaussian Noise&0.188&0.528&0.412&0.359&0.082&0.362&0.311&0.417&0.415&0.228&0.392&0.817\\
&Uniform Noise&0.305&0.532&0.485&0.453&0.131&0.361&0.432&0.512&0.603&0.251&0.492&0.909\\
&Impulse Noise&0.137&0.534&0.425&0.362&0.189&0.364&0.410&0.513&0.451&0.226&0.317&0.909 \\
&Density(L)&-&0.266&-&0.248&-&0.179&-&0.513&-&0.225&-&0.908 \\
&Cutout(L)&-&0.428&-&0.303&-&0.273&-&0.321&-&0.159&-&0.817\\
&Crosstalk(L)&-&0.531&-&0.416&-&0.364&-&0.436&-&0.237&-&0.899 \\
\midrule
\multirow{6}{*}{Object}&Motion Blur(C)&0.375&-&0.483&-&0.524&-&0.745&-&0.494&-&0.521&-\\
&Local Density(L)&-&0.535&-&0.449&-&0.364&-&0.525&-&0.543&-&0.902\\
&Local Cutout(L)&-&0.623&-&0.384&-&0.364&-&0.522&-&0.316&-&0.907\\
&Local Gaussian(L)&-&0.618&-&0.392&-&0.364&-&0.518&-&0.165&-&0.905\\
&Local Uniform(L)&-&0.619&-&0.362&-&0.364&-&0.509&-&0.310&-&0.909\\
&Local Impulse(L)&-&0.532&-&0.401&-&0.364&-&0.419&-&0.299&-&0.908\\
\bottomrule
\end{tabular}
}}
\label{table6}
\end{table*}


\begin{table*}[t]
\small
\caption{The results of different perception tasks on Safety2Drive with digital adversarial attacks.}
\centering
{\resizebox{\linewidth}{!}{
\begin{tabular}{c|c|c|c|c|c|c|c|c|c|c}
\toprule
\multirow{3}{*}{Algo. }&\multirow{3}{*}{Digital Attack} &\multicolumn{8}{c}{Safety2Drive}&\multirow{3}{*}{Avg.}\\
&&\makecell[c]{Sudden Pede. \\ Crossing}&
\makecell[c]{Lane\\ Changing 
}&\makecell[c]{Stationary \\ Obstacle 
}&\makecell[c]{Opposing \\ Passing}&\makecell[c]{Cut In}&\makecell[c]{Cut Out}&\makecell[c]{Turn Right 
}&\makecell[c]{Turn Left}\\
\midrule
\multirow{4}{*}{YoloV5}&Benign&0.883&0.882&0.770&0.839&0.880&0.924&0.990&0.989&0.895\\
&PGD&0.487&0.689&0.598&0.443&0.796&0.651&0.336&0.236&0.471\\
&FGSM&0.243&0.298&0.157&0.215&0.541&0.769&0.207&0.277&0.301\\
\midrule
\multirow{4}{*}{PointPillors}&Benign&0.621&0.296&0.364&0.516&0.905&0.909&0.259&0.805&0.576\\
&PGD&0.197&0.268&0.076&0.132&0.055&0.253&0.063&0.785&0.229\\
&FGSM&0.206&0.251&0.091&0.137&0.119&0.156&0.113&0.799&0.234\\
\midrule
\multirow{4}{*}{CLRNet}&Benign&0.951&0.910&0.913&0.923&0.905&0.952&0.858&0.909&0.915\\
&PGD&0.877&0.827&0.821&0.863&0.898&0.832&0.805&0.784&0.838\\
&FGSM&0.881&0.854&0.794&0.860&0.847&0.565&0.782&0.316&0.737\\
\bottomrule
\end{tabular}
}}
\label{tablescenariodigital}
\end{table*}

\begin{table*}[t]
\small
\caption{The results of object detection  on Safety2Drive with physical adversarial attacks.}
\centering
{\resizebox{\linewidth}{!}{
\begin{tabular}{c|c|c|c|c|c|c|c|c|c|c}
\toprule
\multirow{3}{*}{Algo.}&\multirow{3}{*}{Phy. Attack} &\multicolumn{8}{c}{Safety2Drive}&\multirow{3}{*}{Avg.}\\
&&\makecell[c]{Sudden Pede. \\ Crossing}&
\makecell[c]{Lane\\ Changing 
}&\makecell[c]{Stationary \\ Obstacle 
}&\makecell[c]{Opposing \\ Passing}&\makecell[c]{Cut In}&\makecell[c]{Cut Out}&\makecell[c]{Turn Right 
}&\makecell[c]{Turn Left}\\
\midrule
\multirow{3}{*}{YoloV5}&Benign&0.883&0.882&0.770&0.839&0.880&0.924&0.990&0.989&0.895\\
&Adv.Patch&0.539&0.865&0.513&0.381&0.804&0.850&0.990&0.840&0.723\\
&TC-EGA&0.754&0.727&0.726&0.829&0.668&0.890&0.493&0.851&0.742\\
\bottomrule
\end{tabular}
}}
\label{tablescenariophy}
\end{table*}

\begin{table*}[t]
\small
\caption{The results of object detection  on Safety2Drive with backdoor attack.}
\centering
{\resizebox{\linewidth}{!}{
\begin{tabular}{c|c|c|c|c|c|c|c|c|c|c}
\hline
\multirow{3}{*}{Algo.}&\multirow{3}{*}{Back. Attack} &\multicolumn{8}{c}{Safety2Drive}&\multirow{3}{*}{Avg.}\\
&&\makecell[c]{Sudden Pede. \\ Crossing}&
\makecell[c]{Lane\\ Changing 
}&\makecell[c]{Stationary \\ Obstacle 
}&\makecell[c]{Opposing \\ Passing}&\makecell[c]{Cut In}&\makecell[c]{Cut Out}&\makecell[c]{Turn Right 
}&\makecell[c]{Turn Left}\\
\hline
\multirow{2}{*}{YoloV5}&Benign&0.910&0.986&0.736&0.842&0.782&0.705&0.523&0.990&0.809\\
&OGA Attack&0.676&0.436&0.160&0.546&0.772&0.524&0.297&0.726&0.517\\
\hline
\end{tabular}
}}
\label{tablescenarioback}
\end{table*}
%
\section{Experiments}

We evaluate Safety2Drive using the open-source simulator CARLA with the sensor configurations such as camera, LiDAR, and HD map.
Consequently, Safety2Drive supports various  perception tasks, such as camera-based and LiDAR-based object detection, depth estimation, and lane detection. 
Fig.\ref{perception} shows the prediction results of different perception algorithms in real time on the  \textit{Decelerating} scenario.
We first evaluate the performance of perception module of AD system, and then employ the scenario generalization function to transform the functional test item into safety-critical scenarios. 
On the one hand,  we evaluate the robustness of the camera-based and LiDAR-based object detection on the natural environmental corruption scenarios. 
On the other hand,  we evaluate the safety of perception tasks such as object detection and lane detection on the corresponding adversarial attack scenarios.
Further, we evaluate the functional safety of end-to-end and VLM-based AD systems.  The experiments are executed on a single GeForce RTX 4090D.

In the experiments, we choose 8 functional test items for AD testing: \textit{Sudden Pedestrian Crossing},   \textit{Lane Changing},   \textit{Stationary Obstacle},   \textit{Opposing Passing},   \textit{Cut In},   \textit{Cut Out},   \textit{Turn Right} and \textit{Turn Left}.
Each of test item with parametric configuration corresponds to 10 specific scenarios by adjusting parameters such as the initial state and position of the vehicle,   allowing for batch testing of AD.

\subsection{Evaluation of Natural Environmental Corruption Scenario}
We evaluate the performance of object detection on both standard scenarios and their corresponding natural environment corruption counterparts. In the experiments, the severity is set to 3.
\subsubsection{Baselines and Metric}
For the camera sensor,   we adopt YoloV5 model. For the LiDAR sensor,   the PointPillars~\cite{lang2019pointpillars} model is utilized.
We train YoloV5 and PointPillars using the CARLA training data, and then the pre-trained models are used to evaluate the performance of object detection.
We use AP (0.5) as the evaluation metric,   which is Average Precision (AP) considering an Intersection over Union (IoU) threshold of 0.5,   to evaluate the performance of class ``Car''.
\subsubsection{Results}
The experimental results are shown in Tab.\ref{table6}.
Above all, the performance of object detection model is negatively affected to different degrees when there are various environment corruptions in the driving scenarios.
Specifically, for the camera-based object detection model, snow can have a serious impact over all the scenarios.
For the LiDAR-based object detection model, the impact of environmental corruptions varies across scenarios.
Although environmental corruptions do not have much impact on PointPillars for \textit{Cut Out} scenarios, they lead to performance degradation on most scenarios.
Meanwhile, the impact of environmental corruptions on PointPillars is smaller than that of YoloV5.
\subsection{Evaluation of Adversarial Attack Scenario}
Further, we evaluate the performance of perception tasks such as object detection and lane detection on standard scenarios as well as corresponding adversarial attack scenarios, respectively.

\subsubsection{Baselines and Metric}
For the object detection model we also use camera-based YoloV5 and LiDAR-based PointPillars and report \textit{AP (0.5)} as metric.
For the lane detection we use CLRNet~\cite{zheng2022clrnet}, which is pre-trained on CARLA training data, and report \textit{Accuracy} as metric.
In the adversarial attack scenarios,   for digital attacks,   we use 2 attack algorithms: PGD~\cite{madry2018towards} and FGSM~\cite{goodfellow2014explaining}.
We deploy physical attacks on camera-based object detection task.
Specifically, we employ Adv.Patch~\cite{shrestha2023towards} as a patch-based physical attack and TC-EGA~\cite{hu2022adversarial} as camouflage-based physical attack in the CARLA simulation of physical world.
For backdoor attacks, we use Attack by Aligning~\cite{cheng2023attacking} to attack the object detection model by using Object Generation Attack (OGA) with 30\% posion rate.


\subsubsection{Results}
The experimental results are shown in Tabs.\ref{tablescenariodigital}, \ref{tablescenariophy} and \ref{tablescenarioback}. 
The adversarial attack experiments reveal significant performance degradation in both camera-based and LiDAR-based object detection under ‌digital attack‌ scenarios.
Lane detection model also exhibits compromised robustness in digital attack scenarios.
For ‌physical attacks‌, patch-based and camouflage-based attacks induce measurable performance drops in object detection in most cases, though their efficacy varies by scenario, e.g., patch-based attack notably degrades performance in the \textit{Opposing
Passing} scenario but shows minimal impact in \textit{Turn Right} scenarios, whereas camouflage-based attack demonstrates the opposite trend.
In ‌backdoor attack scenarios, poisoning training data with trigger patterns induces false recognition in object detection tasks during inference.
These experimental results demonstrate that diverse adversarial attack scenarios, i.e., digital, physical, and backdoor, can systematically mislead AD perception systems, posing critical risks to functional safety.
\subsection{Leaderboard of Driving Agent}
In the experiments, we implement some AD systems on Safety2Drive, including both end-to-end and VLM.
We then evaluate these AD systems on the  Leaderboard~\footnote{https://leaderboard.carla.org/evaluation\_v2\_1/} using evaluation metrics.

\subsubsection{Baselines}




(1) TransFuser~\cite{chitta2022transfuser}: a multi-modal fusion transformer that integrates perspective-view RGB and BEV LiDAR representations using self-attention, overcoming the limited contextual reasoning of geometry-only fusion methods. 

(2) VAD~\cite{vad}: a vision-based AD system designed to robustly interpret visual information from the driving environment. VAD focuses on extracting detailed scene context through advanced computer vision algorithms, thereby improving its performance under a variety of driving conditions.

(3) UniAD~\cite{hu2023planning}: a unified AD system that integrates multi-modal sensor data into a single framework. UniAD streamlines the perception,   planning,   and control stages to deliver cohesive decision-making,   particularly under challenging and interactive traffic situations.

(4) LMDrive~\cite{Lmdrive}: an AD system built on a vision-language model. LMDrive not only processes visual inputs but also leverages language-based scene descriptions to enhance its understanding of context and semantics,   ultimately aiming to improve reasoning and overall driving performance.

\subsubsection{Metric}

To evaluate the safety of driving agents, we adopt a set of metrics commonly used in the CARLA leaderboard.
The primary metric is the \textbf{driving score}, denoted as $R$$P$, serving as the product between the \textbf{route completion} $R$ and the \textbf{infractions penalty} $P$.
Specifically, route completion
denotes percentage of the route distance completed by an agent, with a maximum value of 100.
Infraction penalty means that the leaderboard tracks several types of infractions and this metric aggregates all of the ones triggered by an agent.
Each of these infractions has a specific coefficient $c$ based on their severity and is aggregated into the infraction penalty formula:
\begin{equation}
P=\frac{1}{1+\sum_jc_j*\#\text{infractions}_j}
\end{equation}
Infractions related to collisions and failure to comply with traffic regulations, ordered by severity:
\textit{Collisions with pedestrians} ($c=1.0$).
\textit{Collisions with other vehicles} ($c=0.70$).
\textit{Collisions with static elements} ($c=0.60$).
\textit{Running a red light} ($c=0.40$).
\textit{Failure to yield to emergency vehicle} ($c=0.40$).
\textit{Running a stop sign} ($c=0.25$). Agents start with an ideal base score of 1.0 and each infractions reduces it down to a minimum value of 0.



\subsubsection{Results}

\begin{table*}[t]
\small
\caption{Statistics.}
\centering
{\resizebox{\linewidth}{!}{
\begin{tabular}{c|c|cccccccc|c}
\hline
\multirow{3}{*}{Metrics}&\multirow{3}{*}{AD Systems}&\multicolumn{8}{c}{Safety2Drive}&\multirow{3}{*}{Avg.}\\
&&\makecell[c]{Sudden Pedestrian \\ Crossing}&
\makecell[c]{Lane\\ Changing 
}&\makecell[c]{Stationary \\ Obstacle 
}&\makecell[c]{Opposing \\ Passing}&\makecell[c]{Cut In}&\makecell[c]{Cut Out}&\makecell[c]{Turn Right 
}&\makecell[c]{Turn Left}\\
\hline
\multirow{4}{*}{Route Completion}
&TransFuser&67&100&100&46&50.3&100&100&100&\textbf{82.91}\\
&VAD&2.0&96.3&100&43.44&50.12&100&100&100& 73.98 \\
&UniAD&100&94.9&63.3&43.4&48.87&100&100&100 & 81.35\\
&LMDrive&2.0&100&42.6&100&2&100&100&100& 63.33 \\
\hline
\multirow{4}{*}{Infractions Penalty}
&TransFuser&0.5&1.0&1.0&0.65&0.36&0.22&0.32&0.6& 0.58\\
&VAD&1.0&0.65&1.0&0.65&0.6&0.65&0.42&1.0& 0.71 \\
&UniAD&0.65&0.65&0.6&0.65&0.65&0.65&0.45&1.0& 0.66\\
&LMDrive&1.0&1.0&1.0&1.0&1.0&1.0&1.0&1.0& \textbf{1.00} \\
\hline
\multirow{4}{*}{Driving Score}
&TransFuser&33.5&100&100&29.9&18.1&21.6&31.8&60.0& 49.24\\
&VAD&2.0&62.6&100&28.2&30.1&65&42&100&53.7 \\
&UniAD& 65&61.7&38.0&28.2&31.77&65&45.5&100&54.40\\
&LMDrive&2.0 & 100&42.6&100&2.0&100&100&100& \textbf{63.80}\\
\hline
\end{tabular}
}}
\label{tablescenario8}
\end{table*}

Tab.\ref{tablescenario8} reports the results of route completion, infractions penalty and driving score for 4 AD systems on 8 functional test scenarios.
We can see that LMdrive achieves the highest average driving score of 63.80, which has a infractions penalty of 1, even though its route completion metric is not high on the \textit{Sudden Pedestrian
Crossing} and \textit{Cut In} scenarios.
This experiment illustrates the advantage of the VLM AD in preventing traffic violations through scenario understanding over the end-to-end AD models.  In addition, each AD system performs variously for different test items. For example, Garage and UniAD can pass the \textit{Turn Left} scenario, but perform poorly on the \textit{Turn Right} scenario.
This experiments emphasize the necessity of performing different functional tests on AD systems, especially to discover vulnerabilities in the safety of AD systems in certain functions.
\section{Conclusion}
To address the current lack of scenario library for evaluating the functional safety of autonomous driving required by standard regulations in closed-loop simulation, we propose a safety-critical benchmark Safety2Drive. Based on this, we construct a validation framework for autonomous driving, encompassing scenario construction, scenario generalization, and scenario evaluation.
The scenario construction module includes 70 hand-crafted functional test items that meet standard regulatory requirements. The scenario generalization functionality enables the injection of safety-critical risks into the functional test items to assess the safety and robustness of autonomous driving systems.
These safety-critical risks encompass natural environmental corruptions and adversarial attacks across camera and LiDAR sensors. The scenario evaluation module not only supports the evaluation of intelligent perception tasks but also the evaluation of various autonomous driving systems.

\bibliographystyle{named}
\bibliography{neurips25}

\begin{thebibliography}{53}
\providecommand{\natexlab}[1]{#1}
\providecommand{\url}[1]{#1}
\csname url@samestyle\endcsname
\providecommand{\newblock}{\relax}
\providecommand{\bibinfo}[2]{#2}
\providecommand{\BIBentrySTDinterwordspacing}{\spaceskip=0pt\relax}
\providecommand{\BIBentryALTinterwordstretchfactor}{4}
\providecommand{\BIBentryALTinterwordspacing}{\spaceskip=\fontdimen2\font plus
\BIBentryALTinterwordstretchfactor\fontdimen3\font minus \fontdimen4\font\relax}
\providecommand{\BIBforeignlanguage}[2]{{%
\expandafter\ifx\csname l@#1\endcsname\relax
\typeout{** WARNING: IEEEtranN.bst: No hyphenation pattern has been}%
\typeout{** loaded for the language `#1'. Using the pattern for}%
\typeout{** the default language instead.}%
\else
\language=\csname l@#1\endcsname
\fi
#2}}
\providecommand{\BIBdecl}{\relax}
\BIBdecl

\bibitem[Liang et~al.(2020)Liang, Yang, Zeng, Chen, Hu, Casas, and Urtasun]{liang2020pnpnet}
M.~Liang, B.~Yang, W.~Zeng, Y.~Chen, R.~Hu, S.~Casas, and R.~Urtasun, ``Pnpnet: End-to-end perception and prediction with tracking in the loop,'' in \emph{Proceedings of the IEEE/CVF Conference on Computer Vision and Pattern Recognition}, 2020, pp. 11\,553--11\,562.

\bibitem[Jia et~al.(2023)Jia, Wu, Chen, Xie, He, Yan, and Li]{jia2023think}
X.~Jia, P.~Wu, L.~Chen, J.~Xie, C.~He, J.~Yan, and H.~Li, ``Think twice before driving: Towards scalable decoders for end-to-end autonomous driving,'' in \emph{Proceedings of the IEEE/CVF Conference on Computer Vision and Pattern Recognition}, 2023, pp. 21\,983--21\,994.

\bibitem[Chitta et~al.(2022)Chitta, Prakash, Jaeger, Yu, Renz, and Geiger]{chitta2022transfuser}
K.~Chitta, A.~Prakash, B.~Jaeger, Z.~Yu, K.~Renz, and A.~Geiger, ``Transfuser: Imitation with transformer-based sensor fusion for autonomous driving,'' \emph{IEEE transactions on pattern analysis and machine intelligence}, vol.~45, no.~11, pp. 12\,878--12\,895, 2022.

\bibitem[Mei et~al.(2024)Mei, Ma, Yang, Wen, Cai, Li, Fu, Zhang, Cai, Dou, et~al.]{mei2024continuously}
J.~Mei, Y.~Ma, X.~Yang, L.~Wen, X.~Cai, X.~Li, D.~Fu, B.~Zhang, P.~Cai, M.~Dou \emph{et~al.}, ``Continuously learning, adapting, and improving: A dual-process approach to autonomous driving,'' \emph{arXiv preprint arXiv:2405.15324}, 2024.

\bibitem[Kato et~al.(2018)Kato, Tokunaga, Maruyama, Maeda, Hirabayashi, Kitsukawa, Monrroy, Ando, Fujii, and Azumi]{kato2018autoware}
S.~Kato, S.~Tokunaga, Y.~Maruyama, S.~Maeda, M.~Hirabayashi, Y.~Kitsukawa, A.~Monrroy, T.~Ando, Y.~Fujii, and T.~Azumi, ``Autoware on board: Enabling autonomous vehicles with embedded systems,'' in \emph{2018 ACM/IEEE 9th International Conference on Cyber-Physical Systems (ICCPS)}.\hskip 1em plus 0.5em minus 0.4em\relax IEEE, 2018, pp. 287--296.

\bibitem[Kochanthara et~al.(2024)Kochanthara, Singh, Forrai, and Cleophas]{kochanthara2024safety}
S.~Kochanthara, T.~Singh, A.~Forrai, and L.~Cleophas, ``Safety of perception systems for automated driving: A case study on apollo,'' \emph{ACM Transactions on Software Engineering and Methodology}, vol.~33, no.~3, pp. 1--28, 2024.

\bibitem[Liu et~al.(2023)Liu, Tang, Amini, Yang, Mao, Rus, and Han]{liu2023bevfusion}
Z.~Liu, H.~Tang, A.~Amini, X.~Yang, H.~Mao, D.~L. Rus, and S.~Han, ``Bevfusion: Multi-task multi-sensor fusion with unified bird's-eye view representation,'' in \emph{2023 IEEE international conference on robotics and automation (ICRA)}.\hskip 1em plus 0.5em minus 0.4em\relax IEEE, 2023, pp. 2774--2781.

\bibitem[Li et~al.(2024{\natexlab{a}})Li, Jia, Wang, and Yan]{li2024think2drive}
Q.~Li, X.~Jia, S.~Wang, and J.~Yan, ``Think2drive: Efficient reinforcement learning by thinking with latent world model for autonomous driving (in carla-v2),'' in \emph{European Conference on Computer Vision}.\hskip 1em plus 0.5em minus 0.4em\relax Springer, 2024, pp. 142--158.

\bibitem[Renz et~al.(2024)Renz, Chen, Marcu, H{\"u}nermann, Hanotte, Karnsund, Shotton, Arani, and Sinavski]{renz2024carllava}
K.~Renz, L.~Chen, A.-M. Marcu, J.~H{\"u}nermann, B.~Hanotte, A.~Karnsund, J.~Shotton, E.~Arani, and O.~Sinavski, ``Carllava: Vision language models for camera-only closed-loop driving,'' \emph{arXiv preprint arXiv:2406.10165}, 2024.

\bibitem[Wang et~al.(2024)Wang, Yu, Jiang, Lan, Shi, Chang, Kautz, Li, and Alvarez]{wang2024omnidrive}
S.~Wang, Z.~Yu, X.~Jiang, S.~Lan, M.~Shi, N.~Chang, J.~Kautz, Y.~Li, and J.~M. Alvarez, ``Omnidrive: A holistic llm-agent framework for autonomous driving with 3d perception, reasoning and planning,'' \emph{arXiv preprint arXiv:2405.01533}, 2024.

\bibitem[Li et~al.(2022)Li, Chen, Wang, Hong, Ye, Han, Chen, Zhang, Xu, Yeung, et~al.]{li2022coda}
K.~Li, K.~Chen, H.~Wang, L.~Hong, C.~Ye, J.~Han, Y.~Chen, W.~Zhang, C.~Xu, D.-Y. Yeung \emph{et~al.}, ``Coda: A real-world road corner case dataset for object detection in autonomous driving,'' in \emph{European Conference on Computer Vision}.\hskip 1em plus 0.5em minus 0.4em\relax Springer, 2022, pp. 406--423.

\bibitem[Geiger et~al.(2012)Geiger, Lenz, and Urtasun]{geiger2012we}
A.~Geiger, P.~Lenz, and R.~Urtasun, ``Are we ready for autonomous driving? the kitti vision benchmark suite,'' in \emph{2012 IEEE conference on computer vision and pattern recognition}.\hskip 1em plus 0.5em minus 0.4em\relax IEEE, 2012, pp. 3354--3361.

\bibitem[Caesar et~al.(2020)Caesar, Bankiti, Lang, Vora, Liong, Xu, Krishnan, Pan, Baldan, and Beijbom]{caesar2020nuscenes}
H.~Caesar, V.~Bankiti, A.~H. Lang, S.~Vora, V.~E. Liong, Q.~Xu, A.~Krishnan, Y.~Pan, G.~Baldan, and O.~Beijbom, ``nuscenes: A multimodal dataset for autonomous driving,'' in \emph{Proceedings of the IEEE/CVF conference on computer vision and pattern recognition}, 2020, pp. 11\,621--11\,631.

\bibitem[Zhai et~al.(2023)Zhai, Feng, Du, Mao, Liu, Tan, Zhang, Ye, and Wang]{zhai2023rethinking}
J.-T. Zhai, Z.~Feng, J.~Du, Y.~Mao, J.-J. Liu, Z.~Tan, Y.~Zhang, X.~Ye, and J.~Wang, ``Rethinking the open-loop evaluation of end-to-end autonomous driving in nuscenes,'' \emph{arXiv preprint arXiv:2305.10430}, 2023.

\bibitem[Xu et~al.(2022)Xu, Ding, Lyu, Liu, Wang, He, Hu, Zhao, and Li]{xu2022safebench}
C.~Xu, W.~Ding, W.~Lyu, Z.~Liu, S.~Wang, Y.~He, H.~Hu, D.~Zhao, and B.~Li, ``Safebench: A benchmarking platform for safety evaluation of autonomous vehicles,'' \emph{Advances in Neural Information Processing Systems}, vol.~35, pp. 25\,667--25\,682, 2022.

\bibitem[Jia et~al.(2024)Jia, Yang, Li, Zhang, and Yan]{jiabench2drive}
X.~Jia, Z.~Yang, Q.~Li, Z.~Zhang, and J.~Yan, ``Bench2drive: Towards multi-ability benchmarking of closed-loop end-to-end autonomous driving,'' in \emph{The Thirty-eight Conference on Neural Information Processing Systems Datasets and Benchmarks Track}, 2024.

\bibitem[Zhang et~al.(2024{\natexlab{a}})Zhang, Xu, and Li]{zhang2024chatscene}
J.~Zhang, C.~Xu, and B.~Li, ``Chatscene: Knowledge-enabled safety-critical scenario generation for autonomous vehicles,'' in \emph{Proceedings of the IEEE/CVF Conference on Computer Vision and Pattern Recognition}, 2024, pp. 15\,459--15\,469.

\bibitem[Li et~al.(2024{\natexlab{b}})Li, Qin, and Widdershoven]{li2024iss}
R.~Li, T.~Qin, and C.~Widdershoven, ``Iss-scenario: Scenario-based testing in carla,'' in \emph{International Symposium on Theoretical Aspects of Software Engineering}.\hskip 1em plus 0.5em minus 0.4em\relax Springer, 2024, pp. 279--286.

\bibitem[Dong et~al.(2023)Dong, Kang, Zhang, Zhu, Wang, Yang, Su, Wei, and Zhu]{dong2023benchmarking}
Y.~Dong, C.~Kang, J.~Zhang, Z.~Zhu, Y.~Wang, X.~Yang, H.~Su, X.~Wei, and J.~Zhu, ``Benchmarking robustness of 3d object detection to common corruptions,'' in \emph{Proceedings of the IEEE/CVF Conference on Computer Vision and Pattern Recognition}, 2023, pp. 1022--1032.

\bibitem[Wang et~al.(20245)Wang, Xie, Sato, Luo, Xu, and Chen]{wang2024revisiting}
N.~Wang, S.~Xie, T.~Sato, Y.~Luo, K.~Xu, and Q.~A. Chen, ``Revisiting physical-world adversarial attack on traffic sign recognition: A commercial systems perspective,'' in \emph{ISOC Network and Distributed System Security Symposium (NDSS)}, 20245.

\bibitem[Zhang et~al.(2024{\natexlab{b}})Zhang, Wang, Li, Xiao, Liang, Liu, Liu, and Tao]{zhang2024lanevil}
T.~Zhang, L.~Wang, H.~Li, Y.~Xiao, S.~Liang, A.~Liu, X.~Liu, and D.~Tao, ``Lanevil: Benchmarking the robustness of lane detection to environmental illusions,'' in \emph{Proceedings of the 32nd ACM International Conference on Multimedia}, 2024, pp. 5403--5412.

\bibitem[Kothari et~al.(2021)Kothari, Kreiss, and Alahi]{kothari2021human}
P.~Kothari, S.~Kreiss, and A.~Alahi, ``Human trajectory forecasting in crowds: A deep learning perspective,'' \emph{IEEE Transactions on Intelligent Transportation Systems}, vol.~23, no.~7, pp. 7386--7400, 2021.

\bibitem[Sun et~al.(2020)Sun, Kretzschmar, Dotiwalla, Chouard, Patnaik, Tsui, Guo, Zhou, Chai, Caine, et~al.]{sun2020scalability}
P.~Sun, H.~Kretzschmar, X.~Dotiwalla, A.~Chouard, V.~Patnaik, P.~Tsui, J.~Guo, Y.~Zhou, Y.~Chai, B.~Caine \emph{et~al.}, ``Scalability in perception for autonomous driving: Waymo open dataset,'' in \emph{Proceedings of the IEEE/CVF conference on computer vision and pattern recognition}, 2020, pp. 2446--2454.

\bibitem[Feng et~al.(2023)Feng, Sun, Yan, Zhu, Zou, Shen, and Liu]{feng2023dense}
S.~Feng, H.~Sun, X.~Yan, H.~Zhu, Z.~Zou, S.~Shen, and H.~X. Liu, ``Dense reinforcement learning for safety validation of autonomous vehicles,'' \emph{Nature}, vol. 615, no. 7953, pp. 620--627, 2023.

\bibitem[Hao et~al.(2023)Hao, Cui, Luo, Xie, Bai, Yang, Yan, Pan, and Yang]{hao2023adversarial}
K.~Hao, W.~Cui, Y.~Luo, L.~Xie, Y.~Bai, J.~Yang, S.~Yan, Y.~Pan, and Z.~Yang, ``Adversarial safety-critical scenario generation using naturalistic human driving priors,'' \emph{IEEE Transactions on Intelligent Vehicles}, 2023.

\bibitem[Althoff et~al.(2017)Althoff, Koschi, and Manzinger]{althoff2017commonroad}
M.~Althoff, M.~Koschi, and S.~Manzinger, ``Commonroad: Composable benchmarks for motion planning on roads,'' in \emph{2017 IEEE Intelligent Vehicles Symposium (IV)}.\hskip 1em plus 0.5em minus 0.4em\relax IEEE, 2017, pp. 719--726.

\bibitem[Shen et~al.(2022)Shen, Wang, Wan, Luo, Sato, Hu, Zhang, Guo, Zhong, Li, et~al.]{shen2022sok}
J.~Shen, N.~Wang, Z.~Wan, Y.~Luo, T.~Sato, Z.~Hu, X.~Zhang, S.~Guo, Z.~Zhong, K.~Li \emph{et~al.}, ``Sok: On the semantic ai security in autonomous driving,'' \emph{arXiv preprint arXiv:2203.05314}, 2022.

\bibitem[Wang et~al.(2021{\natexlab{a}})Wang, Yao, Liu, Li, Hao, and Zhu]{wang2021can}
W.~Wang, Y.~Yao, X.~Liu, X.~Li, P.~Hao, and T.~Zhu, ``I can see the light: Attacks on autonomous vehicles using invisible lights,'' in \emph{Proceedings of the 2021 ACM SIGSAC Conference on Computer and Communications Security}, 2021, pp. 1930--1944.

\bibitem[Zhang et~al.(2024{\natexlab{c}})Zhang, Liu, Zhang, Liang, and Liu]{zhang2024towards}
X.~Zhang, A.~Liu, T.~Zhang, S.~Liang, and X.~Liu, ``Towards robust physical-world backdoor attacks on lane detection,'' in \emph{Proceedings of the 32nd ACM International Conference on Multimedia}, 2024, pp. 5131--5140.

\bibitem[Cao et~al.(2022)Cao, Xiao, Anandkumar, Xu, and Pavone]{cao2022advdo}
Y.~Cao, C.~Xiao, A.~Anandkumar, D.~Xu, and M.~Pavone, ``Advdo: Realistic adversarial attacks for trajectory prediction,'' in \emph{European Conference on Computer Vision}.\hskip 1em plus 0.5em minus 0.4em\relax Springer, 2022, pp. 36--52.

\bibitem[Li et~al.(2024{\natexlab{c}})Li, Qiu, Wang, Xu, Singh, Yamazaki, Chen, Huang, and Raj]{li2024r}
X.~Li, K.~Qiu, J.~Wang, X.~Xu, R.~Singh, K.~Yamazaki, H.~Chen, X.~Huang, and B.~Raj, ``R 2-bench: Benchmarking the robustness of referring perception models under perturbations,'' in \emph{European Conference on Computer Vision}.\hskip 1em plus 0.5em minus 0.4em\relax Springer, 2024, pp. 211--230.

\bibitem[Hu et~al.(2022{\natexlab{a}})Hu, Shen, Guo, Zhang, Zhong, Chen, and Li]{hu2022pass}
Z.~Hu, J.~Shen, S.~Guo, X.~Zhang, Z.~Zhong, Q.~A. Chen, and K.~Li, ``Pass: A system-driven evaluation platform for autonomous driving safety and security,'' in \emph{NDSS Workshop on Automotive and Autonomous Vehicle Security (AutoSec)}, 2022.

\bibitem[Beemelmanns et~al.(2024)Beemelmanns, Zhang, Geller, and Eckstein]{beemelmanns2024multicorrupt}
T.~Beemelmanns, Q.~Zhang, C.~Geller, and L.~Eckstein, ``Multicorrupt: A multi-modal robustness dataset and benchmark of lidar-camera fusion for 3d object detection,'' in \emph{2024 IEEE Intelligent Vehicles Symposium (IV)}.\hskip 1em plus 0.5em minus 0.4em\relax IEEE, 2024, pp. 3255--3261.

\bibitem[Xie et~al.(2025)Xie, Kong, Zhang, Ren, Pan, Chen, and Liu]{xie2025benchmarking}
S.~Xie, L.~Kong, W.~Zhang, J.~Ren, L.~Pan, K.~Chen, and Z.~Liu, ``Benchmarking and improving bird's eye view perception robustness in autonomous driving,'' \emph{IEEE Transactions on Pattern Analysis and Machine Intelligence}, 2025.

\bibitem[Goodfellow et~al.(2014)Goodfellow, Shlens, and Szegedy]{goodfellow2014explaining}
I.~J. Goodfellow, J.~Shlens, and C.~Szegedy, ``Explaining and harnessing adversarial examples,'' \emph{arXiv preprint arXiv:1412.6572}, 2014.

\bibitem[Madry et~al.(2018)Madry, Makelov, Schmidt, Tsipras, and Vladu]{madry2018towards}
A.~Madry, A.~Makelov, L.~Schmidt, D.~Tsipras, and A.~Vladu, ``Towards deep learning models resistant to adversarial attacks,'' in \emph{International Conference on Learning Representations}, 2018.

\bibitem[Brown et~al.(2017)Brown, Man{\'e}, Roy, Abadi, and Gilmer]{brown2017adversarial}
T.~B. Brown, D.~Man{\'e}, A.~Roy, M.~Abadi, and J.~Gilmer, ``Adversarial patch,'' \emph{arXiv preprint arXiv:1712.09665}, 2017.

\bibitem[Liu et~al.(2024)Liu, Wu, Wang, Han, Guo, Xiang, and Zhang]{liu2024beware}
H.~Liu, Z.~Wu, H.~Wang, X.~Han, S.~Guo, T.~Xiang, and T.~Zhang, ``Beware of road markings: A new adversarial patch attack to monocular depth estimation,'' \emph{Advances in Neural Information Processing Systems}, vol.~37, pp. 67\,689--67\,711, 2024.

\bibitem[Athalye et~al.(2018)Athalye, Engstrom, Ilyas, and Kwok]{athalye2018synthesizing}
A.~Athalye, L.~Engstrom, A.~Ilyas, and K.~Kwok, ``Synthesizing robust adversarial examples,'' in \emph{International conference on machine learning}.\hskip 1em plus 0.5em minus 0.4em\relax PMLR, 2018, pp. 284--293.

\bibitem[Sharif et~al.(2016)Sharif, Bhagavatula, Bauer, and Reiter]{sharif2016accessorize}
M.~Sharif, S.~Bhagavatula, L.~Bauer, and M.~K. Reiter, ``Accessorize to a crime: Real and stealthy attacks on state-of-the-art face recognition,'' in \emph{Proceedings of the 2016 acm sigsac conference on computer and communications security}, 2016, pp. 1528--1540.

\bibitem[Mahendran and Vedaldi(2015)]{mahendran2015understanding}
A.~Mahendran and A.~Vedaldi, ``Understanding deep image representations by inverting them,'' in \emph{Proceedings of the IEEE conference on computer vision and pattern recognition}, 2015, pp. 5188--5196.

\bibitem[Shrestha et~al.(2023)Shrestha, Pathak, and Viegas]{shrestha2023towards}
S.~Shrestha, S.~Pathak, and E.~K. Viegas, ``Towards a robust adversarial patch attack against unmanned aerial vehicles object detection,'' in \emph{2023 IEEE/RSJ International Conference on Intelligent Robots and Systems (IROS)}.\hskip 1em plus 0.5em minus 0.4em\relax IEEE, 2023, pp. 3256--3263.

\bibitem[Wang et~al.(2021{\natexlab{b}})Wang, Liu, Yin, Liu, Tang, and Liu]{wang2021dual}
J.~Wang, A.~Liu, Z.~Yin, S.~Liu, S.~Tang, and X.~Liu, ``Dual attention suppression attack: Generate adversarial camouflage in physical world,'' in \emph{Proceedings of the IEEE/CVF conference on computer vision and pattern recognition}, 2021, pp. 8565--8574.

\bibitem[Zhang et~al.(2023)Zhang, Gong, Zhang, Bin, Li, Qi, Wen, and Zhong]{zhang2023boosting}
Y.~Zhang, Z.~Gong, Y.~Zhang, K.~Bin, Y.~Li, J.~Qi, H.~Wen, and P.~Zhong, ``Boosting transferability of physical attack against detectors by redistributing separable attention,'' \emph{Pattern Recognition}, vol. 138, p. 109435, 2023.

\bibitem[Zhang et~al.(2024{\natexlab{d}})Zhang, Pan, Liu, Yan, Choo, and Wang]{zhang2024backdoor}
S.~Zhang, Y.~Pan, Q.~Liu, Z.~Yan, K.-K.~R. Choo, and G.~Wang, ``Backdoor attacks and defenses targeting multi-domain ai models: A comprehensive review,'' \emph{ACM Computing Surveys}, vol.~57, no.~4, pp. 1--35, 2024.

\bibitem[Pourkeshavarz et~al.(2024)Pourkeshavarz, Sabokrou, and Rasouli]{pourkeshavarz2024adversarial}
M.~Pourkeshavarz, M.~Sabokrou, and A.~Rasouli, ``Adversarial backdoor attack by naturalistic data poisoning on trajectory prediction in autonomous driving,'' in \emph{Proceedings of the IEEE/CVF Conference on Computer Vision and Pattern Recognition}, 2024, pp. 14\,885--14\,894.

\bibitem[Lang et~al.(2019)Lang, Vora, Caesar, Zhou, Yang, and Beijbom]{lang2019pointpillars}
A.~H. Lang, S.~Vora, H.~Caesar, L.~Zhou, J.~Yang, and O.~Beijbom, ``Pointpillars: Fast encoders for object detection from point clouds,'' in \emph{Proceedings of the IEEE/CVF conference on computer vision and pattern recognition}, 2019, pp. 12\,697--12\,705.

\bibitem[Zheng et~al.(2022)Zheng, Huang, Liu, Tang, Yang, Cai, and He]{zheng2022clrnet}
T.~Zheng, Y.~Huang, Y.~Liu, W.~Tang, Z.~Yang, D.~Cai, and X.~He, ``Clrnet: Cross layer refinement network for lane detection,'' in \emph{Proceedings of the IEEE/CVF conference on computer vision and pattern recognition}, 2022, pp. 898--907.

\bibitem[Hu et~al.(2022{\natexlab{b}})Hu, Huang, Zhu, Sun, Zhang, and Hu]{hu2022adversarial}
Z.~Hu, S.~Huang, X.~Zhu, F.~Sun, B.~Zhang, and X.~Hu, ``Adversarial texture for fooling person detectors in the physical world,'' in \emph{Proceedings of the IEEE/CVF conference on computer vision and pattern recognition}, 2022, pp. 13\,307--13\,316.

\bibitem[Cheng et~al.(2023)Cheng, Hu, and Cheng]{cheng2023attacking}
Y.~Cheng, W.~Hu, and M.~Cheng, ``Attacking by aligning: Clean-label backdoor attacks on object detection,'' \emph{arXiv preprint arXiv:2307.10487}, 2023.

\bibitem[Jiang et~al.(2023)Jiang, Chen, Xu, Liao, Chen, Zhou, Zhang, Liu, Huang, and Wang]{vad}
B.~Jiang, S.~Chen, Q.~Xu, B.~Liao, J.~Chen, H.~Zhou, Q.~Zhang, W.~Liu, C.~Huang, and X.~Wang, ``Vad: Vectorized scene representation for efficient autonomous driving,'' in \emph{Proceedings of the IEEE/CVF International Conference on Computer Vision}, 2023, pp. 8340--8350.

\bibitem[Hu et~al.(2023)Hu, Yang, Chen, Li, Sima, Zhu, Chai, Du, Lin, Wang, et~al.]{hu2023planning}
Y.~Hu, J.~Yang, L.~Chen, K.~Li, C.~Sima, X.~Zhu, S.~Chai, S.~Du, T.~Lin, W.~Wang \emph{et~al.}, ``Planning-oriented autonomous driving,'' in \emph{Proceedings of the IEEE/CVF conference on computer vision and pattern recognition}, 2023, pp. 17\,853--17\,862.

\bibitem[Shao et~al.(2024)Shao, Hu, Wang, Waslander, Liu, and Li]{Lmdrive}
H.~Shao, Y.~Hu, L.~Wang, S.~L. Waslander, Y.~Liu, and H.~Li, ``Lmdrive: Closed-loop end-to-end driving with large language models,'' in \emph{Proceedings of the IEEE/CVF conference on computer vision and pattern recognition}, 2024, pp. 15\,120--15\,130.

\end{thebibliography}








\appendix

\section{Functional Test Items}

In Tab.~\ref{tab2}, we summary 6 main categories covering 70 standardized regulatory-compliant functional test items.
\begin{table}[h]
\small
\caption{Main categories and functional test items.}
\centering
{\resizebox{0.99\linewidth}{!}{
\begin{tabular}{c|p{13cm}}
\toprule
Main test categories&Functional test items\\
\midrule
Adaptive Cruise Control&StraightRoadCruising,   CurvedRoadCruising,  DownhillCruising,  StraightRoadLaneChangeLeft,  CurvedRoadLaneDepartureLeft,  CurvedRoadLaneDepartureRight,  StraightRoadLaneDepartureLeft,  StraightRoadLaneDepartureRight\\
\midrule
Following Driving&StationaryTargetVehicleStraightRoad,   LowSpeedTargetVehicleStraightRoad,   DeceleratingTargetVehicleStraightRoad,   TargetVehicleCutOutStraightRoad,   MisidentifiedOvertakingStraightRoad,   SingleTrafficParticipantStraightRoad,   MultipleTrafficParticipants,   StraightRoadLaneChangeLeftWithDeceleratingLead,   VehicleEntryDetectionAndResponse,   BicycleRidingAlongRoad,   StableCarFollowing,   StopAndGoFunction,   StraightRoadMixedSlowVehicles,   StraightRoadPedestrianAndVehicleSlow,   CurvedRoadPedestrianAndVehicleSlow,   BicycleCutOut\\
\midrule
Emergency Braking&HighSpeedCutInStraightRoad,   PostCutOutLeadVehicleStraightRoad,   PedestrianCrossingRoad,   BicycleCrossingRoad,  StraightVehicleConflict,   RightTurnVehicleConflict,  LeftTurnVehicleConflict,  RoundaboutNavigation,  CurvedRoadLeadDeceleration,  CrosswalkDetectionWithPedestrian,  CurvedRoadLeadEmergencyBrake,  NightRainStraightRoadTruckEmergencyBrake,  OppositeLaneInvasionDetection,  CurvedRoadMixedSlowVehicles,  BicycleCutIn,  BicycleCutOutWithStaticPedestrian,  LeadBicycleDeceleration\\
\midrule
Traffic Sign&SpeedLimitSignRecognitionAndResponse, StopYieldSignRecognitionAndResponse, LaneMarkingRecognitionAndResponse, CrosswalkRecognitionAndResponse, TrafficLightRecognitionAndResponse, DirectionalSignalRecognitionAndResponse, SpeedLimitActivationAndDeactivation, CurvedRoadSpeedLimit \\
\midrule
Overtaking&PedestrianObstacleDetection, PedestrianWalkingAlongRoad, Overtaking, StraightRoadPostCutOutStaticCar, StraightRoadCutInAndStop, CurvedRoadStaticMotorcycleAndCar, CurvedRoadStaticPedestrianAndCar, StraightRoadCarAccident, CurvedRoadAccidentWithPedestrianAndCar, DayRainStraightRoadCutOutWithCones, TrafficConeDetection, StreetObstacleDetection, AccidentWarningObjectDetection, BicycleCutOutWithMovingPedestrian \\
\midrule
Parking&EmergencyRoadsideParking, RightmostLaneParking, ParkingSpotRecognition \\
\midrule
Merging&AdjacentLaneMergeWithoutVehicles, AdjacentLaneMergeWithVehicles, LaneChangeHighwayEntranceRecognition, HighwayExitLeadVehicleDetection\\
\bottomrule
\end{tabular}
}}
\label{tab2}
\end{table}
\section{LLM-based Scenario Generation}
\begin{figure*}[h]
  \centering
\includegraphics[width=0.48\linewidth,  scale=1]{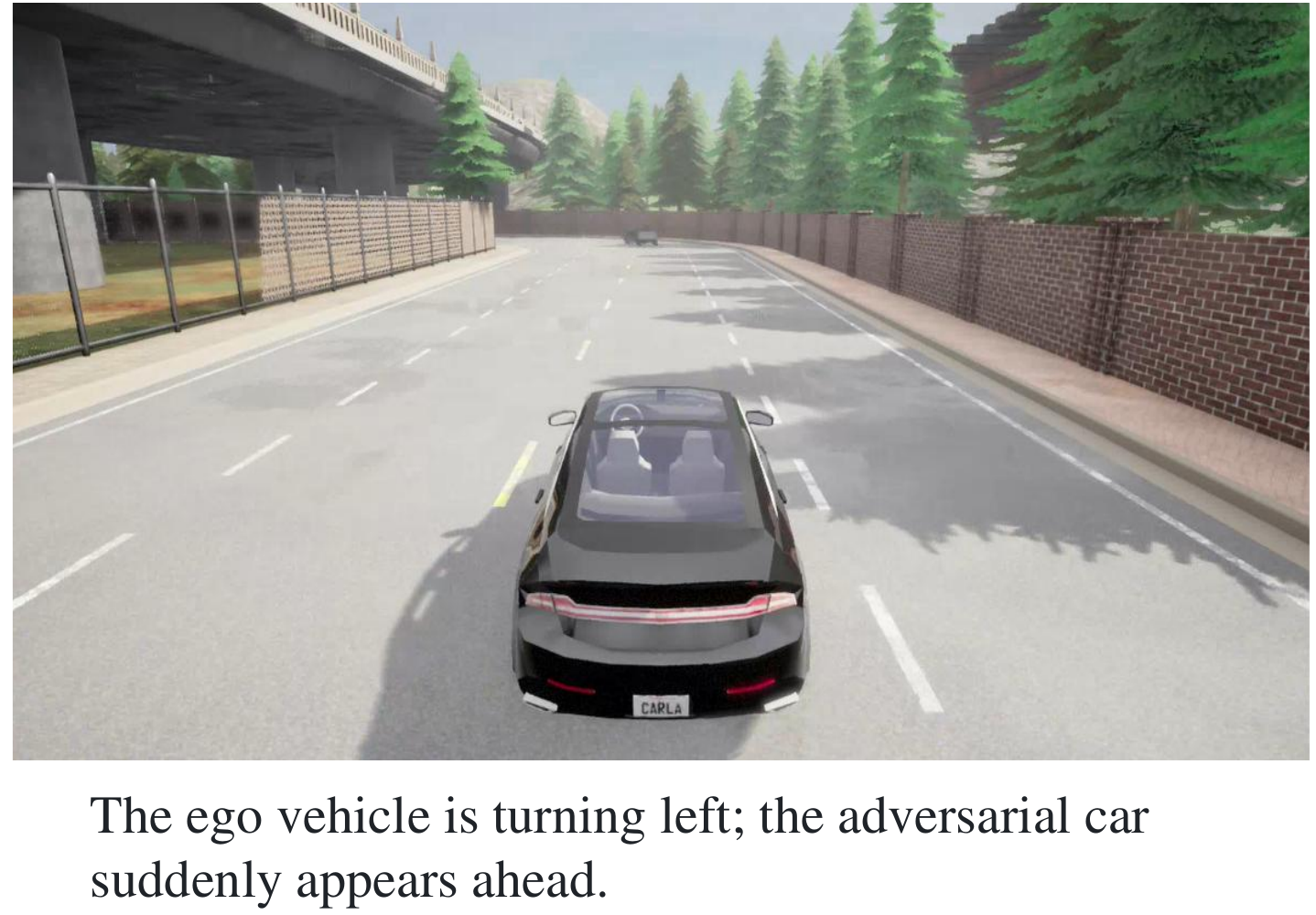}
\includegraphics[width=0.48\linewidth,  scale=1]{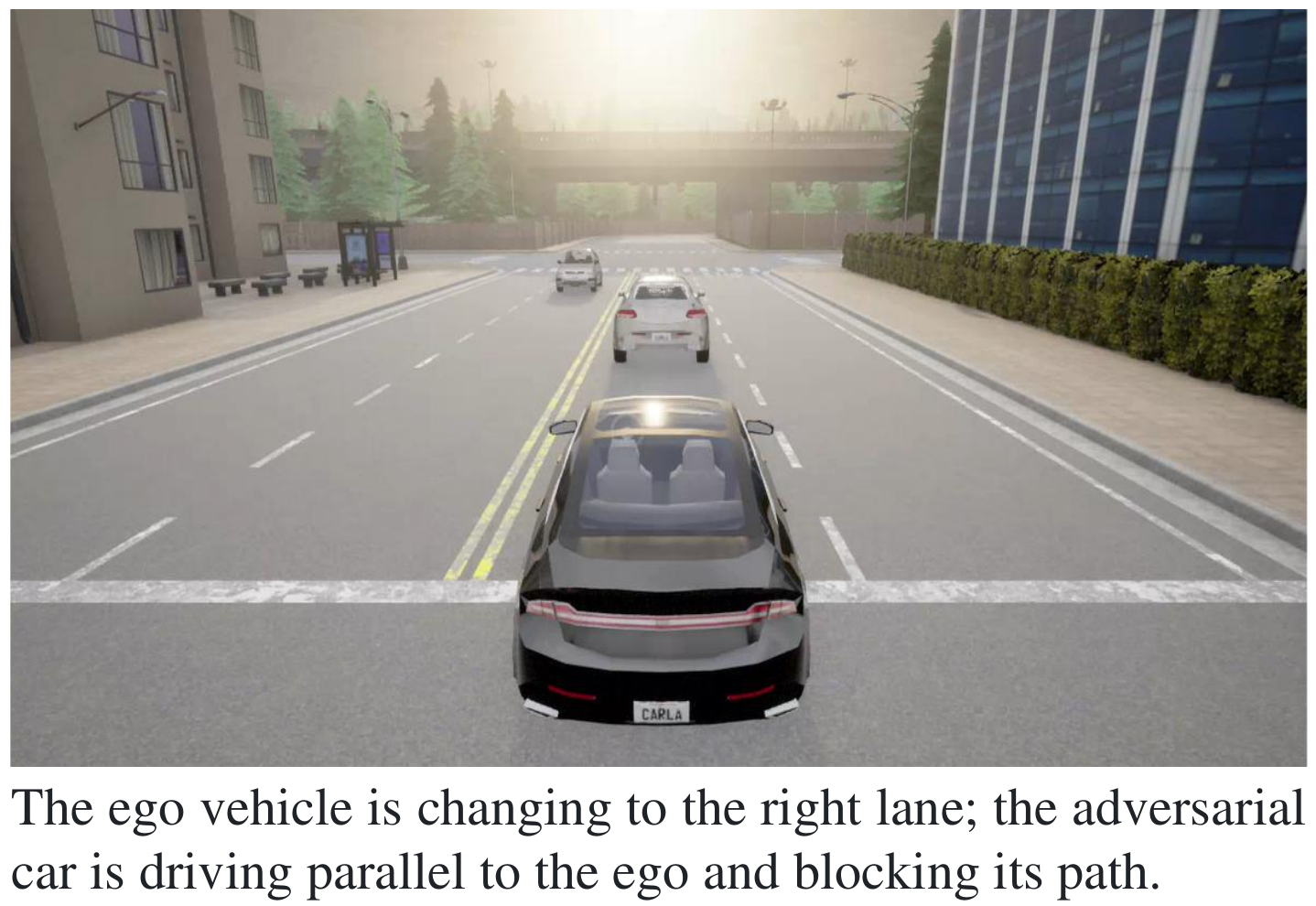}
  \caption{Two cases of LLM-based scenario generation algorithm using ChatScene.}
  \label{demo}
\end{figure*}

%
The Safety2Drive supports LLM-based scenario generation, which is used as an auxiliary capability for the flexible and fast construction. We integrate the  open source LLM-based scenario generation algorithm ChatScene. The ChatScene receives user input as a prompt and then generates executable scripts to simulate the scenarios in the CARLA simulation environment.
Some examples of text-to-scenario
mappings are provided in Fig.~\ref{demo},
\section{More Results of Natural Environment Corruptions}
\begin{figure*}[h]
  \centering
\includegraphics[width=0.98\linewidth,  scale=1]{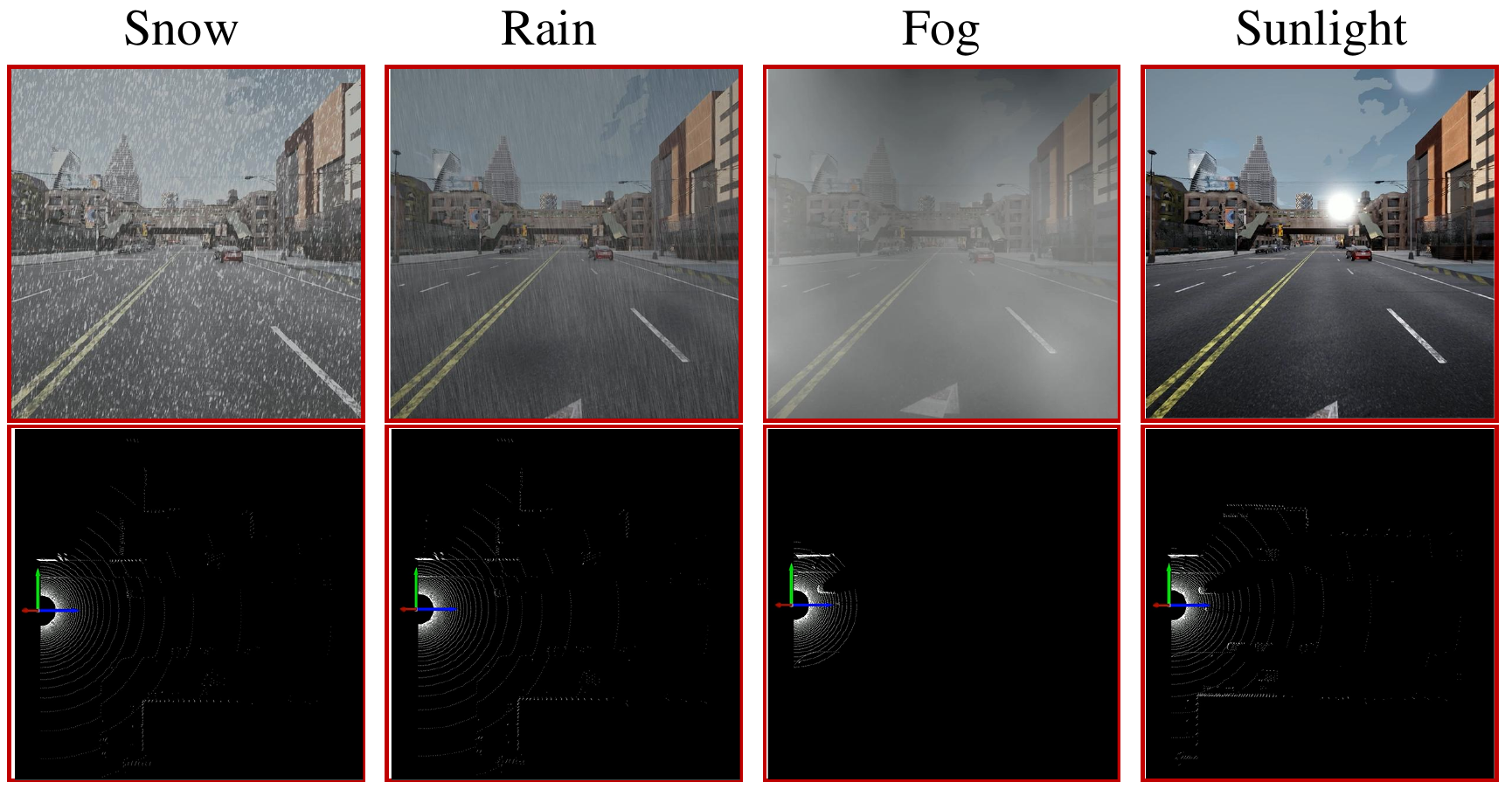}
  \caption{ The first part of the visualization results of all corruptions. The images or point clouds in red boxes are modified under the corresponding corruption, while the images or point clouds in black boxes are kept unchanged.}
  \label{vis1}
\end{figure*}
\begin{figure*}[h]
  \centering
\includegraphics[width=0.98\linewidth,  scale=1]{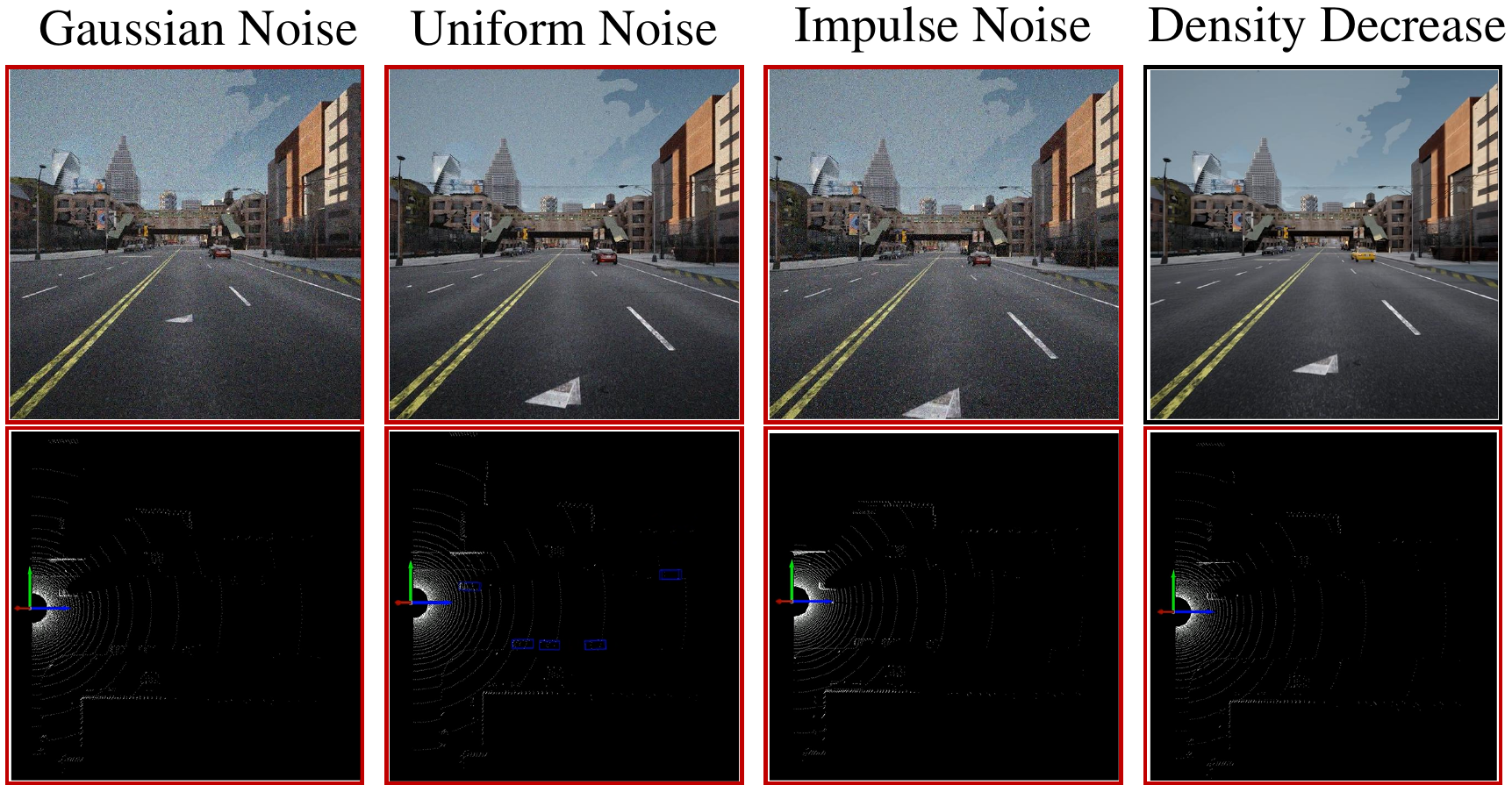}
  \caption{ The second part of the visualization results of all corruptions.}
  \label{vis2}
\end{figure*}
\begin{figure*}[h]
  \centering
\includegraphics[width=0.98\linewidth,  scale=1]{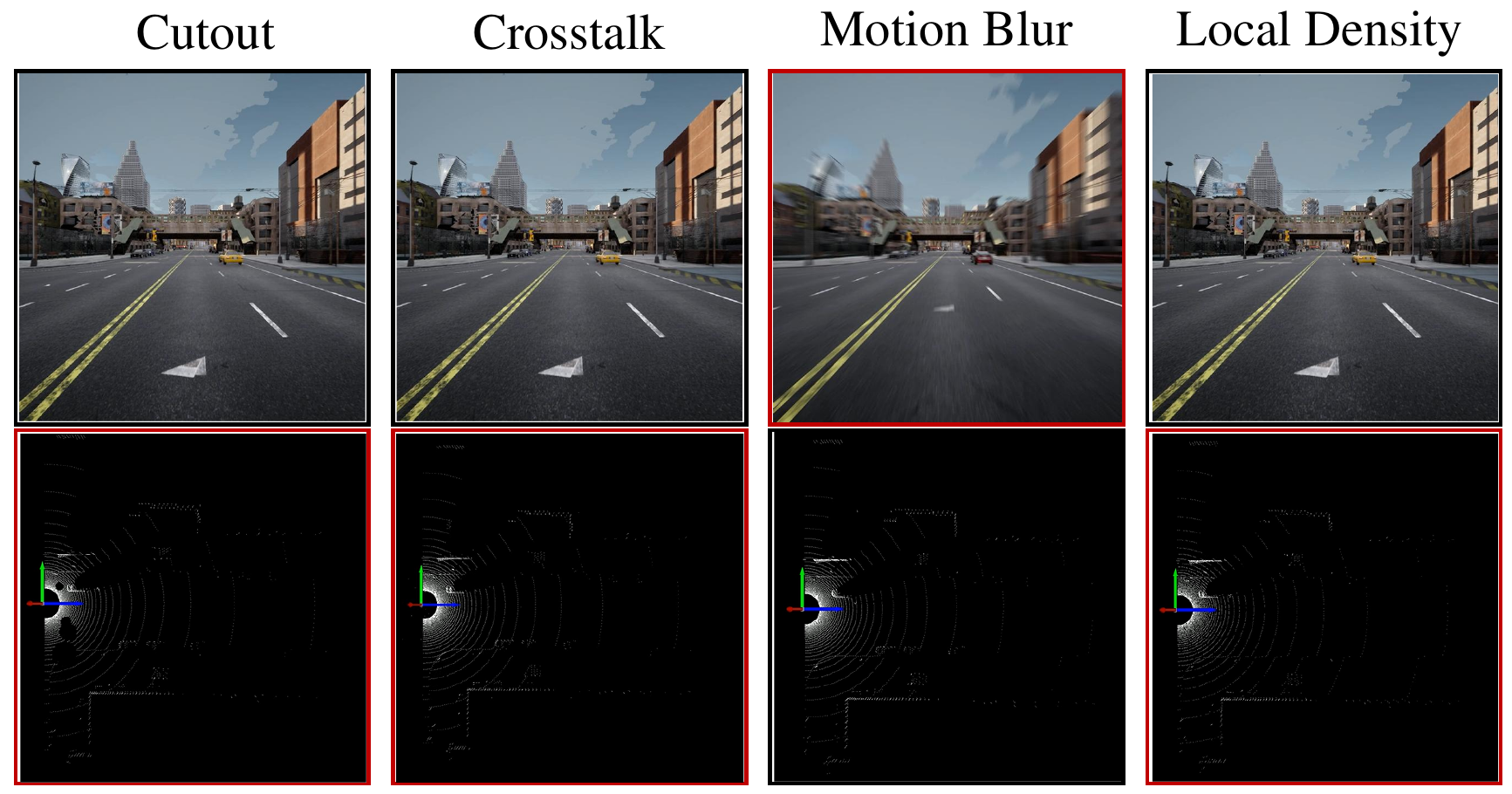}
  \caption{ The third part of the visualization results of all corruptions.}
  \label{vis3}
\end{figure*}
\begin{figure*}[h]
  \centering
\includegraphics[width=0.98\linewidth,  scale=1]{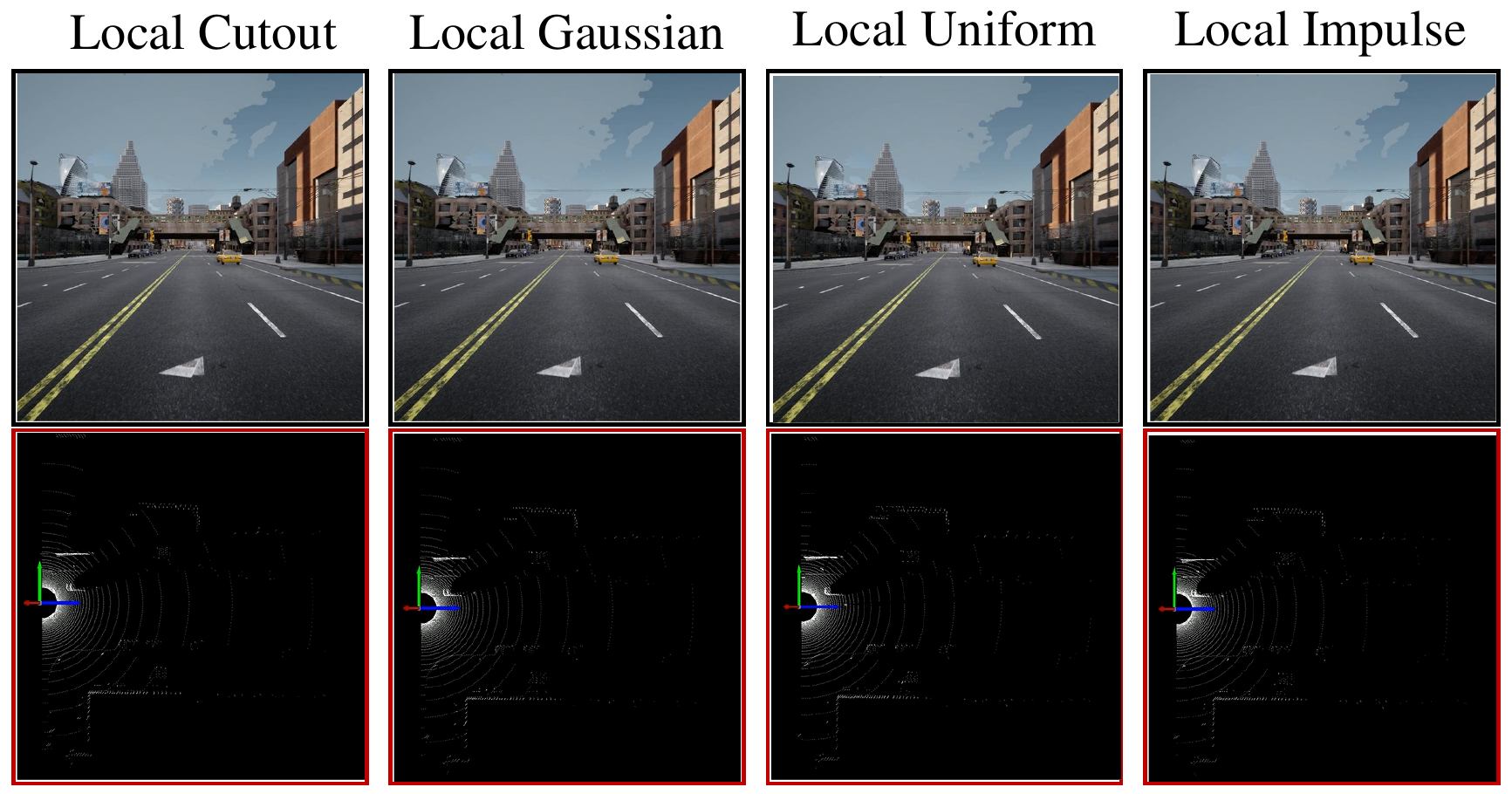}
  \caption{ The fourth part of the visualization results of all corruptions.}
  \label{vis4}
\end{figure*}
We show the visualization of all 16 natural environment corruptions in Figs.\ref{vis1}, \ref{vis2}, \ref{vis3} and \ref{vis4}.
Note that an input (image or point cloud) may not be modified under a corruption, thus we mark it by the black box. 
For input that has been modified under the corruption, we mark it by the red box.

\end{document}